\documentclass[11pt]{article}
\usepackage{verbatim,amsmath,amssymb}
\usepackage{epsfig,float,color}
\usepackage{epstopdf}
\usepackage{geometry}
\usepackage{setspace}
\usepackage{wrapfig}
\usepackage{hyperref}
\usepackage[utf8]{inputenc}
\usepackage{graphicx}
\usepackage{flushend}
\usepackage[linesnumbered,ruled]{algorithm2e}

\usepackage[font={footnotesize}]{caption}
\usepackage[font={footnotesize}]{subcaption}
\usepackage{footnote}
\usepackage{multicol}
\usepackage{multirow}
\usepackage{enumitem}   
\usepackage{soul}
\usepackage{framed}
\usepackage[titletoc,title]{appendix}
\usepackage{bm}
\usepackage{siunitx}
\usepackage{cite}
\usepackage[makeroom]{cancel} 
\usepackage{hyperref}
\definecolor{darkblue}{rgb}{0,0,1}
\hypersetup{pdftex=true, colorlinks=true, breaklinks=true, linkcolor=magenta, menucolor=magenta, citecolor=magenta, urlcolor=magenta}

\newcommand{\PKadd}[1]{\textcolor{black}{#1}}
\def\BibTeX{{\rm B\kern-.05em{\sc i\kern-.025em b}\kern-.08em
		T\kern-.1667em\lower.7ex\hbox{E}\kern-.125emX}}
\usepackage{balance}
\usepackage{hyperref}
\definecolor{beaublue}{rgb}{0.94, 0.97, 1.0}
\definecolor{darkblue}{rgb}{0, 0, 1}
\definecolor{LightCyan}{rgb}{0.88,1,1}
\definecolor{mygreen}{RGB}{28,172,0}
\definecolor{mylilas}{RGB}{170,55,241}
\definecolor{grayblue}{RGB}{220,230,240}
\definecolor{topressbg}{rgb}{0.95,0.95,0.92}
\definecolor{topressgreen}{rgb}{0,0.5,0}
\definecolor{topressgray}{rgb}{0.5,0.5,0.5}
\definecolor{topresspurple}{rgb}{0.58,0,0.82}
\definecolor{topressblue}{rgb}{0,0,1}
\hypersetup{
	pdftex=true,
	colorlinks=true,
	breaklinks=true,
	linkcolor=darkblue,
	menucolor=darkblue,
	pagecolor=darkblue,
	citecolor=darkblue,
	urlcolor=darkblue
}
\newcommand{\ms}[1]{\mathrm{#1}} %
\newcommand{\mb}[1]{\mathbf{#1}} 
\newcommand{\trr}[1]{{#1}^{\!\top}}
\newcommand{\inv}[1]{{#1}^{\text{-}1}}
\usepackage{pgfplots}
\usetikzlibrary{positioning, arrows.meta}
\tikzset{%
	myarrow/.style = {-Stealth, shorten >=5pt}
}

\newenvironment{rcases}
{\left.\begin{aligned}}
	{\end{aligned}\right\rbrace}

\definecolor{darkblue}{rgb}{0,0,1}
\hypersetup{pdftex=true, colorlinks=true, breaklinks=true, linkcolor=darkblue, menucolor=darkblue, pagecolor=darkblue, citecolor=darkblue, urlcolor=darkblue}

\usepackage{array}
\usepackage{mwe}
\newcolumntype{C}[1]{>{\centering\arraybackslash}m{#1}}

\begin{document}
	
	\begin{center}
		\Large{\bf{Soft Pneumatic Grippers: Topology optimization, 3D-printing and Experimental validation}}\\
		
	\end{center}
	
	\begin{center}
		Prabhat Kumar$^{*,}$$\footnote{pkumar@mae.iith.ac.in}$, Chandra Prakash$^*$, Josh Pinskier$^\dagger$, David Howard$^\dagger$, Matthijs Langelaar$^\ddagger$
		
		\vspace{4mm}
		\small{$*$\textit{Department Mechanical and Aerospace Engineering. India Institute of Technology Hyderbad, Telangana 502285, India}}\\
		\small{$\dagger$\textit{CSIRO Robotics, Pullenvale, QLD 4069, Australia}}\\
		\small{$\ddagger$\textit{Faculty of Mechanical Engineering, Delft University of Technology, Mekelweg 2, Delft, 2628CD, Zuid-Holland, The Netherlands}}
		
		Published\footnote{This pdf is the personal version of an article whose final publication is available at \href{https://www.sciencedirect.com/science/article/pii/S0094114X26001813}{https://www.sciencedirect.com/journal/mechanism-and-machine-theory}} 
		in \textit{Mechanism and Machine Theory }, 
		\href{https://www.sciencedirect.com/science/article/pii/S0094114X26001813?via%3Dihub}{DOI:10.1016/j.mechmachtheory.2026.106531} \\
		Submitted on 18.~April 2026, Accepted on 13. June 2026
	\end{center}
	
	\vspace{3mm}
	\rule{\linewidth}{.15mm}
	{\bf Abstract:}
Typically, heuristic/trial-based approaches are used to design soft pneumatic grippers (SPGs). This paper presents a systematic topology optimization framework for developing SPGs. The design-dependent nature of actuating load is modeled using Darcy's law with an added drainage term. A 2D soft arm unit is then optimized as a compliant mechanism under pneumatic loading. To ensure the design is robust and manufacturable, the problem is formulated as a min-max optimization, where output deformations of blueprint and eroded designs are considered. A volume constraint is imposed on the blueprint part, while a strain-energy constraint is enforced on the eroded part. The Method of Moving Asymptotes is employed to solve optimization problems. The optimized 2D part is extruded suitably to generate a 3D unit. Ten such 3D units are assembled to create a gripper arm.  Both the optimized 2D unit and the corresponding gripper arm outperform their conventional rectangular designs under pneumatic loading, demonstrating the efficacy of the proposed approach. The arms are fabricated using the SLA printing technique. Numerical and experimental results are compared at different pneumatic loads. Four 3D-printed arms are integrated with a supporting structure to form the SPG. The gripping action of the SPG is demonstrated on objects with different weights, sizes, structures, stiffnesses, and shapes.  \\
	

	\vspace{-4mm}
	\rule{\linewidth}{.15mm}
	
  \section{Introduction}
  The use of soft grippers is steadily growing, especially for handling delicate and fragile objects. Because they are made from flexible materials (with a Young's modulus in the kilopascal to megapascal range), their grip causes little to no damage to the objects they hold~\cite{shintake2018soft,zhang2018design,xavier2022soft,kumar2022towards}. In addition, their inherent flexibility and adaptability make them well-suited for sensitive, constrained, and unstructured environments, as well as human-friendly~\cite{kumar2022towards,pinskier2021bioinspiration}. Further, these grippers deliver a high power-to-weight ratio. They are used in many applications including constant force output~\cite{liu2020optimal,liu2021topology,reddy2023topology}, food packaging/handling~\cite{wang2017prestressed,wang2020dual}, underwater activities~\cite{galloway2016soft,mura2018soft}, bio-medical~\cite{rateni2015design,kumar2021topologyBio,hermoza20253d}, to name a few. While these grippers can be powered in various ways~\cite{zhang2018design,xavier2022soft,pinskier2021bioinspiration,kumar2022towards,wang2017prestressed,wang2020dual,chen2018topology,liu2018optimalsoro}, one common type is the soft pneumatic grippe (SPG), which is driven by pneumatic/air pressure loads~\cite{zhang2018design,xavier2022soft,kumar2022towards}. SPGs contain internal air chambers or networks (PneuNets)~\cite{mosadegh2014pneumatic,kumar2022towards} that bend or twist when pressurized. This motion creates an effective grip, making them highly popular for grasping delicate items like fruits, vegetables, and eggs etc.~\cite{pinskier2021bioinspiration,kumar2022towards}. Despite their growing demand and versatility, SPGs are generally designed through trial-and-error, relying heavily on a designer's personal experience and expertise~\cite{kumar2022towards}. In addition,  these design methods often have to be restarted from scratch for different gripping tasks, which increases costs and limits scalability. To solve these issues, this work introduces a \textit{systematic design framework using topology optimization} to automate SPG design process. A distinct attribute of the topology optimization approach is that it generates the optimized solution automatically without knowing the topology of the design a priori. A 2D pneumatic chamber is optimized under design-dependent pressure load. The SPG arm, constructed from the optimized design, is fabricated via SLA 3D printing. With a developed experimental setup, the performance of the SPG is demonstrated through grasping experiments on objects of various shapes, sizes, structures, stiffnesses, and weights.
  
  Topology optimization (TO), a computational design approach, provides an optimized material layout within the design domain to achieve a specific objective while adhering to given constraints. The domain is parametrized using standard/advanced finite elements (FEs)~\cite{kumar2022honeytop90}. Each element is typically assigned a design variable $\rho$, which is assumed constant within that element. During the optimization process, $\rho$ toggles between `0' and `1', where $\rho =1$ and $\rho=0$ indicate the solid and void material states, respectively. Ideally, the final optimized design should consist of FEs with $\rho = 1$, indicating a precise solid-void distribution. However, due to the relaxed nature of TO formulation~\cite{bendsoe2013topology}, FEs with $0 < \rho < 1$ also exist in the solution~\cite{bendsoe2013topology}.
  
  In typical applications, the applied loads are constant, and therefore \textit{design-independent}; however, in the case of  SPGs, actuation is driven by pneumatic loads whose point of action and direction evolve with TO iterations, making them inherently \textit{design-dependent}~\cite{kumar2022topological}. Furthermore, the characteristics of a soft robot are similar to those of a compliant mechanism (CM), as both utilize their flexible members to perform their respective tasks \cite{kumar2022towards,pinskier2024diversity,lu2022optimal,zhang2018design,chi2025topology}. Therefore, a repeating part of the SPG/PneuNet unit can be designed as a CM actuated by a design-dependent pneumatic load~\cite{kumar2022towards,kumar2022topological}. The design-dependent nature of these loads significantly influences the resulting topology of the mechanism or soft robot~\cite{kumar2020topology,kumar2024sorotop}. However, such loads introduce several distinct challenges within a TO framework~\cite{kumar2020topology,kumar2023topress}, as their direction, location, and magnitude change during the optimization process. These challenges  become even more pronounced in compliant mechanism optimization; as a result, to date, only a few methods have been reported for designing pressure-actuated CMs (Pa-CMs)~\cite{kumar2020topology,kumar2022topological}.
  
  Panganiban {\rm et al.}~\cite{panganiban2010topology} provide a nonconforming finite element method to design Pa-CMs. de Souza and Silva~\cite{de2020topology} employ the mix-formulation method. Lu and Tong~\cite{lu2022optimal,lu2021topology} use the moving isosurface method to design an SPG and Pa-CMs.  Chi and Liu~\cite{chi2025topology} employ a multi-criteria approach, considering the design-dependent nature of the pneumatic load. Zhang {\rm et al.}~\cite{zhang2018design} optimize an SPG while maximizing the output deformation for bending motion. A skeleton TO method is proposed by Chen {\rm et al.}~\cite{chen2021enhancing}. Liu {\rm et al.}~\cite{liu2022topology} optimize an SPG while maximizing mutual potential energy. Though design-dependent nature of the load is important, has not been considered in Refs.~\cite{zhang2018design,chen2021enhancing,liu2022topology}. Kumar {\rm et al.}~\cite{kumar2020topology} present a method using Darcy's law with a drainage term for designing Pa-CMs. The approach uses standard FE method and automatically calculates load sensitivities using the adjoint-variable method. This framework has been successfully applied to design 3D Pa-CMs~\cite{kumar2020topology3Dpressure}, robust pressure actuated CMs~\cite{kumar2022topological}, monolithic multi-material grippers~\cite{pinskier2024diversity}, etc.  Therefore, we adopt the Darcy-based approach to model the design-dependent pneumatic loads for optimizing SPG's PneuNet.
  
  Although soft pneumatic grippers undergo large deformations during actuation, linear topology optimization (LTO) has been widely and successfully employed in prior studies on soft pneumatic grippers/soft pneumatic bending actuators/ PneuNet-based systems, for example, see Refs.~\cite{zhang2018design,zhang2018topology,chen2019optimal,pinskier2021bioinspiration,guo2021design,chen2018topology,liu2018optimalsoro,pinskier2024diversity,lu2022optimal,chi2025topology,liu2022topology}. In these works, LTO is primarily used to determine the optimized material distribution and structural layout that governs the deformation mode. At the same time, the final designs are validated through nonlinear finite element analyses, including both material and geometric nonlinearities. Additionally, designs derived from the optimization are successfully fabricated and tested for the intended applications. Given previous studies and the demonstrated efficacy of the optimized designs, we use LTO herein to achieve the optimized unit for the proposed SPG, while not claiming optimality, considering its nonlinear deformation performance.
  
  To ensure the effectiveness and reliability of the design optimization process, the performance of fabricated SPGs must closely match that of the numerical simulations. With the gradient-based optimizer, conventional density-based TO provides optimized designs with gray elements ($0<\rho<1$). In addition, the final CMs are prone to having point connections~\cite{kumar2022topological,wang2011projection}. Extracting such optimized designs can alter the final designs, thereby affecting their performance. To circumvent these issues,  we employ the robust formulation with a blueprint and eroded fields to achieve an optimized unit of SPGs. The optimized blueprint design is smoothened and extracted to create a corresponding 2D CAD model. Nonlinear finite element analyses are performed on both the optimized and conventional pressure chambers using ABAQUS for a comparative study. The 2D CAD models are then extruded and patterned sequentially to achieve respective 3D gripper arms. Numerical simulations and experimentation are conducted at various pressure loadsto demonstrate and compare the bending behavior of the optimized SPG arm. Four arms are fabricated using SLA 3D printing, and an experimental setup is developed. Finally, the experiments are conducted to demonstrate the grasping action on various objects. Within the context of the SPG design, this paper offers the following new contributions:
  \begin{itemize}
  	\item It proposes a robust topology optimization approach to optimize SPGs while explicitly considering the design-dependent nature of the actuating load, which is modeled using Darcy's law with an additional drainage term
  	\item It optimizes a 2D PneuNet unit using linear topology optimization, treating it as a compliant mechanism design under pneumatic actuation, and demonstrates that the optimized unit outperforms a conventional rectangular design.
  	\item It creates a 3D arm by appropriately extruding and patterning the 2D-optimized PneuNet unit, provides a comparative study of nonlinear finite element and experimental deformations of  SPG's arm at different pneumatic loads and demonstrates that the soft pneumatic arm constructed with the optimized chambers outperforms the one with conventional chambers.
  	\item It employs the SLA printing technique to fabricate gripper arms for experimental purposes, and conducts a comparative study between the 3D-printed prototype and its numerical counterpart, revealing, by and large, a good overall agreement between the two.
  	\item It develops a customized experimental setup and demonstrates the versatile grasping performance of the assembled gripper on various objects of different shapes, sizes, structures, stiffnesses, and weights.
  \end{itemize}

  The remainder of the paper is structured as follows. Sec.~\ref{Sec:TOframe} describes pneumatic load modeling,  TO formulation, and sensitivity analysis. Sec.~\ref{Sec:OptimizedUnit} provides a 2D optimized unit and finite element analyses comparison with the conventional chamber. In Sec.~\ref{Sec:3DSPG}, 3D SPG is designed. FE analyses are performed on a soft arm under different pressure loading conditions. SPG is 3D printed. Experiments are performed to grip objects of various shapes, sizes, and weights. Lastly, closing remarks are presented in Sec.~\ref{Sec:Con}.
  
 \section{Topology optimization framework}\label{Sec:TOframe}
\begin{figure}[H]
	\centering
	\begin{subfigure}{0.30\textwidth}
		\includegraphics[scale=.80]{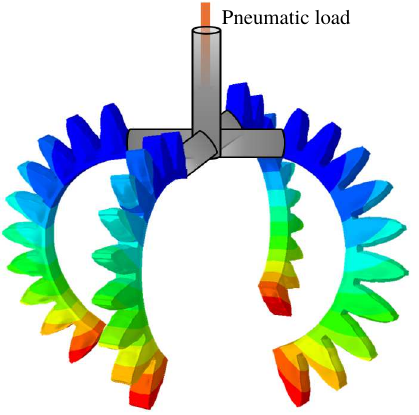}
		\caption{}
	\end{subfigure}
	\qquad 
	\begin{subfigure}{0.30\textwidth}
		\includegraphics[scale=1.0]{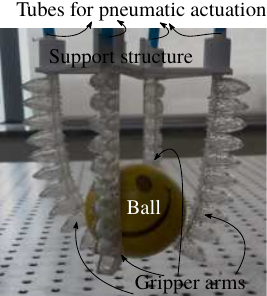}
		\caption{}
	\end{subfigure}
	\quad
	\begin{subfigure}{0.30\textwidth}
		\includegraphics[scale=0.90]{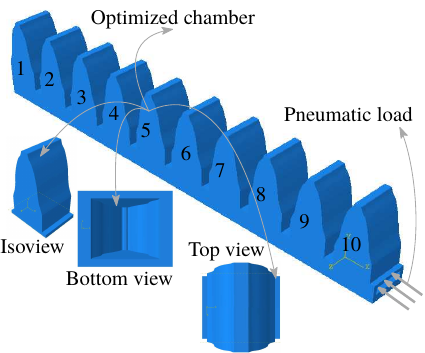}
		\caption{}\label{fig:3D_CAD_opt}
	\end{subfigure}
	\caption{(a) Numerically simulated gripping action  (Sec.~\ref{Sec:3DSPG}) (b) Functional demonstration of a 3D-printed SPG for gripping of a ball (Sec.~\ref{Sec:3DSPG}) (c) 3D CAD model of one arm of the SPG with optimized pressure chambers. A SPG arm is constituted by 10 optimized pressure chambers. Iso-, bottom-, and top-view images of one of the optimized pressure chambers are also depicted.} \label{fig:SPG_four_def}
\end{figure}

\begin{figure}
	\centering
	\includegraphics[scale=0.750]{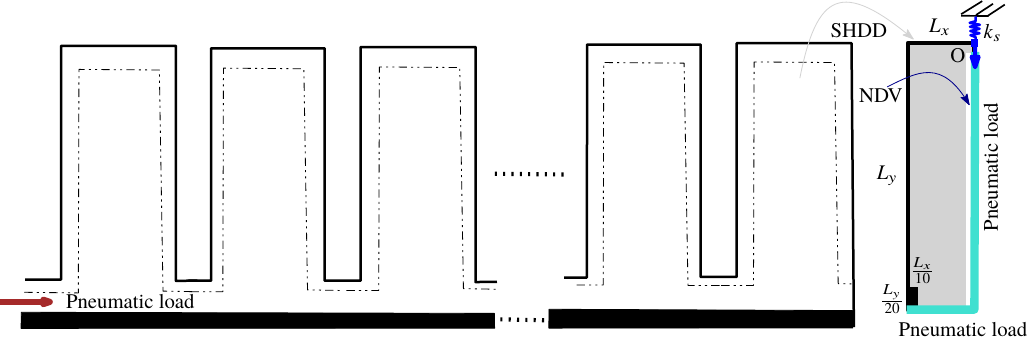}
	\caption{(a) A 2D schematic diagram for the SPG arm. The symmetric half-domain design (SHDD) is shown on the left, with pneumatic and fixed boundary conditions. $k_{s}$ indicate the output spring stiffness. $L_x$ and $L_y$ denote dimensions in $x$ and $y$ directions, respectively. Non-design solid (NDS) and non-design void (NDV) regions are also shown. NDs and NDV regions do not change with optimization iterations.}  \label{fig:DD}
\end{figure}

This section provides the TO framework for the proposed approach. The designed soft gripper consists of four flexible, beam-like arms\footnote{We call arm(s) henceforth in the paper.} capable of bending to aid grasping, as shown in Fig.~\ref{fig:SPG_four_def}. Each compliant beam/arm is composed of 10 units/ PneuNets. The modular configuration of the arm offers scalability, adaptability, and compliance for the gripping action. The focus of this work is to optimize the geometry of a single 2D PneuNet unit using LTO, which, when assembled in series as 3D units\footnote{A 2D optimized unit is appropriately extruded to form a corresponding 3D unit}, forms a complete compliant arm (Fig.~\ref{fig:3D_CAD_opt}). The idea is that, upon actuation with pneumatic pressure, these optimized arms collectively enable effective gripping (Fig.~\ref{fig:SPG_four_def}). The 2D  design domain used for analysis and optimization to determine the optimized material layout is depicted in  Fig.~\ref{fig:DD}.  Note that although considering the entire gripper design domain would expand the design search space, it would introduce additional challenges, including complex actuation modeling, intricate pressure inlet design, complicated boundary conditions, high computational cost, and manufacturability considerations, among others. Thus, the objective of the current manuscript is to optimize the internal geometry of a 2D PneuNet unit chamber for the intended SPG design.

We use the density-based TO method~\cite{kumar2022honeytop90}. The fixed reference domain is parameterized using the quadrilateral FEs wherein each element is assigned a design variable, $\rho$, which is assumed to be constant within that element. The robust formulation that includes three variable fields (design, filtered and projected)--$\bm{\rho},\,\tilde{\bm{\rho}},\,\bar{\bm{\rho}}$ is employed~\cite{wang2011projection}.  The filtered variable $\tilde{\rho}_i$ for element $i$ is defined as~\cite{bruns2001topology}
\begin{equation}\label{eq:densityfilter}
	\tilde{\rho}_i = \frac{\displaystyle\sum_{j \in n_i}v_j\rho_j w(\bm{x}_{i,\,j})}{\displaystyle\sum_{j \in n_i} v_j w(\bm{x}_{i,\,j})}, \quad  	w(\bm{x}_{i,\,j}) =  \max\left(0,\, 1- \frac{||\bm{x}_i^c - \bm{x}_j^c||}{r_\text{min}}\right),
\end{equation}
where $n_i = \{i,\,||\mb{x}_i^c - \mb{x}_j^c||\le r_{\ms{\min}}\}$, indicates the number of neighboring elements for element~$i$ determined as per filter radius, $r_{\min}$.  $\bm{x}_i^c\,\text{and}\, \bm{x}_j^c$ represent center coordinates of element $i$ and element $j$, respectively. $||\,.\,||$ denotes distance in the Euclidean space. Volume of element~$i$ is denoted by $v_i$. $w(\bm{x}_{i,\,j})$ is a linearly decaying weighting function determined as above.  The derivative of $\tilde{\rho}_i$ with respect to $\rho_j$ is determined as
\begin{equation}\label{eq:der_filter}
	\frac{\partial \tilde{\rho_i}}{\partial \rho_j} = \frac{v_j w(\bm{x}_{i,\,j})}{\sum_{k \in n_i}v_k w(\bm{x}_{k,\,j})},
\end{equation}
$\bar{\rho}_i$ is the physical/projected design variable of element $i$, which is determined as~\cite{wang2011projection}
\begin{equation}\label{eq:Heaviside_fn}
	\bar{\rho}_i =  \mathcal{H} \left(\tilde{\rho_e},\eta_p,\beta_p\right)= \frac{\tanh\left(\eta_p\beta_p \right) + \tanh \left(\beta_p(\tilde{\rho}_i -\eta_p \right)}{\tanh \left(\eta_p\beta_p \right) + \tanh \left(\beta_p(1-\eta)\right)},
\end{equation}
where $\eta$ defines the threshold of the projection and  $\beta_p\in [1,\,\infty)$ indicates the steepness of projection function. Ideally, $\beta_p\to\infty$, for achieving discrete or close-to-binary ($0-1$) solutions. Practically, however, $\beta_p$ is increased from its initial value, $\beta_\text{int}=1$, to a user defined maximum value $\beta_\text{max}$ using a continuation scheme. The derivative of $\bar{\rho}_i$ with respect to $ \rho_i$ is given as
\begin{equation}\label{eq:derHeavy}
	\frac{\partial \bar{\rho_i}}{\partial \tilde{\rho}_i} = \beta_p \frac{1-\tan\left(\beta_p\left(\tilde{\rho_i}-\eta_p\right)\right)^2}{\tanh \left(\eta_p\beta_p \right) + \tanh \left(\beta_p(1-\eta_p)\right)}.
\end{equation}
Elastic modulus of each element is interpolated using the modified SIMP formulation as
\begin{equation}
	E_i = E_0 + \bar{\rho}_i^p(E_1-E_0),
\end{equation} 
where $E_1$ and $E_0$ are elastic modulii of solid and void phases of the element, respectively. The material contrast $\frac{E_0}{E_1} = 10^{-9}$ is used. $p$ indicates the SIMP penalty parameter and is set to 3 for the optimization problem solved in Sec.~\ref{Sec:OptimizedUnit}. Next, we describe the design-dependent pneumatic/fluid pressure load modeling.

\subsection{Pneumatic load modeling}
As mentioned before, a pneumatic load is inherently design-dependent and this characteristic must be included while modeling such loads within a TO setting~\cite{kumar2020topology}. We use the Darcy approach~\cite{kumar2020topology,kumar2023topress} that provides an easy and efficient way to handle the design-dependent behavior of the pneumatic loads that is, it efficiently accounts for the current pressure distribution as topology evolves. The approach leverages evolving material states of the FEs by initially treating them as a porous media, whereas the provided pneumatic loading conditions furnishes pressure gradient for the reference domain~\cite{kumar2022topological,kumar2020topology,kumar2023topress,kumar2020topology3Dpressure}. Mathematically, the Darcy flux is determined as~\cite{kumar2020topology}
\begin{equation}\label{eq:Darcy_flux}
	q =  -\frac{\kappa}{\mu}\nabla p = -\mathcal{K} (\bar{\bm{\rho}})\nabla p,
\end{equation}
where $\nabla p$, $\kappa$ and $\mu$ indicate the pressure gradient, permeability 
the medium and fluid viscosity, respectively.  $\mathcal{K} (\bar{\bm{\rho}})$, the flow coefficient, depends upon the porous media, that is,  in a TO setting, it relies upon the material states of the element. For element $i$, mathematically it can be written as~\cite{kumar2020topology,kumar2020topology3Dpressure}
\begin{equation}
	\mathcal{K}_i = \mathcal{K}_v\left(1-\left(1-\epsilon\right) \mathcal{H} \left(\tilde{\rho_i},\eta_k,\beta_k\right) \right),
\end{equation}
where flow contrast, $\epsilon = \frac{\mathcal{K}_s}{\mathcal{K}_v} = 10^{-7}$ is used~\cite{kumar2020topology3Dpressure,kumar2023topress}. The flow coefficients for void and solid phases are respresented by $\mathcal{K}_v$ and $\mathcal{K}_s$, respectively. $\left\{\eta_k,\,\beta_k\right\}$ are the flow parameters, where the first term defines the step and second indicates slope of $\mathcal{K}(\bar{\rho_i})$~\cite{kumar2020topology,kumar2023topress}. $\mathcal{K}_v =1$ and $\epsilon = 1\times10^{-7}$are set, i.e., $\mathcal{K}_s = 1\times10^{-7}$. It is noted in~\cite{kumar2020topology,kumar2020topology3Dpressure,kumar2023topress} that pressure field obtained from  Eq.~\ref{eq:Darcy_flux} fails to indicate the natural pressure variation in a TO setting; thus, a volumetric drainage term $Q_\text{drain}$ is added to circumvent this and obtain the expected pressure field~\cite{kumar2020topology,kumar2020topology3Dpressure,kumar2023topress}.  One evaluates
\begin{equation}\label{eq:Drainterm}
	Q_\text{drain} =-\mathbb{D} \left(p-p_0\right),
\end{equation}
where $\mathbb{D} = \mathbb{D}_\text{s} \mathcal{H} (\bar{\rho_i},\eta_\text{d},\beta_\text{d})$, and  $\left\{\eta_\text{d},\,\beta_\text{d}\right\}$ analogous to $\left\{\eta_k,\,\beta_k\right\}$,  are the drainage parameters. $\mathbb{D}_\text{s} = \frac{\ln r}{\Delta s} \mathcal{K}_\text{s}$, where $r \in [0.001,\, 0.1] = \frac{p|_{\Delta s}}{p_\text{in}}$. To reduce the number of parameters within the Darcy formulation, $\left\{\eta_k,\,\beta_k\right\} = \left\{\eta_\text{d},\,\beta_\text{d}\right\} = \left\{\eta_f,\,\beta_f\right\} $ is considered~\cite{kumar2023topress,kumar2024sorotop}. $\Delta s$ is a penetration parameter, and $p|_{\Delta s}$ is the pressure at $\Delta s$. The balance equation with $Q_\text{drain}$ becomes~\cite{kumar2020topology}
\begin{equation}\label{eq:balance_eq_with_Q}
	\nabla\cdot\bm{q} -Q_\text{drain} = 0.
\end{equation}
In context of the standard finite element formulation, Eq.~\ref{eq:balance_eq_with_Q} transpires to~\cite{kumar2020topology}
\begin{equation}\label{eq:PDEsolutionpressure}
	\mathbf{Ap} = \mathbf{0},
\end{equation}
where $\mathbf{A}$ and $\mathbf{p}$ are the global flow matrix and pressure vector, respectively. By solving Eq.~\ref{eq:PDEsolutionpressure}, one obtains the pressure field, which is converted to the nodal loads using the following equation as~\cite{kumar2020topology}
\begin{equation}
	\mathbf{F} = -\mathbf{T} \mathbf{p},
\end{equation}
where $\mathbf{F}$ is the global force vector and $\mathbf{T}$ is the the global transformation vector~\cite{kumar2020topology,kumar2023topress}.
 
\subsection{Optimization formulation}
The characteristics of a PneuNet, monolithic design with no rigid links, are similar to a Pa-CM as mentioned before~\cite{kumar2022towards}. The physical variables corresponding to blueprint and eroded designs are obtained using  $0.5 + \Delta \eta$ and $0.5$ instead of $\eta$ in Eq.~\ref{eq:Heaviside_fn}. We minimize the maximum output deformations obtained for the blueprint and eroded designs with linear mechanics assumptions. The volume constraint is applied to the blueprint design, whereas a strain energy constraint is applied to the eroded design~\cite{kumar2024sorotop}. The latter constraint is applied so the optimized PneuNet/unit can be sustained under the applied pneumatic load. Mathematically, 
\begin{equation}\label{eq:Optimizationequation}
	\begin{rcases}
		\underset{\bar{\bm{\rho}}(\tilde{\bm{\rho}}(\bm{\rho}))}{\text{min}:}
		f_{\text{unit}} = \max \left\{u^{o}_b,\,u^{o}_e\right\} = \max\left\{\bm{l}^\top \mathbf{u}_b,\,\bm{l}^\top \mathbf{u}_e\right\}\\
		\text{such that:} \\
		\,{^u}{\bm{\lambda}_q}:\,\, \mathbf{A}_q\mathbf{p}_q = \mathbf{0 },\, |_{q =e,\,b}\\
		\,{^p}\bm{\lambda}_q:\,\,  \mathbf{K}_q\mathbf{u}_q = \mathbf{F}_q = -\mathbf{T} \mathbf{p}_q\\
		\,{\Lambda}_b:\,\, \text{g}_1 = {V_b}/{V^*}\le 1\\
		\,{\Lambda}_e:\,\,  \text{g}_2= {se_e}/{se^*} \le 1\\
		\, 0\le \rho_i,\,\tilde{\rho}_i,\,\bar{\rho}_i\le 1 
	\end{rcases},
\end{equation}
where the qualities related to the eroded and blueprint are indicated with $e$ and $b$ subscripts. $u_q^o$ indicate the output deformation in the desired direction (Fig.~\ref{fig:DD}). $\bm{l}$ is a vector with all zeros except the entry corresponding to the output degree of freedom set to one. 	$\mathbf{K}_q$ and $\mathbf{u}_q$ denote the global stiffness matrix and global displacement vector, respectively. g$_1$ is a volume constraint, where $V^*$ and $V_b$ provide the blueprint design's permitted and current volume fractions, respectively. Constraint g$_2$ is a strain energy constraint with  $se_e$ and $se^*$ as the eroded design's current and defined strain energy, respectively. The latter is evaluated at the beginning of the optimization loop as  $se^* = S_f\times se_e$, where $se_e = \frac{1}{2}\mathbf{u}_e^\top\mathbf{K}_e \mathbf{u}_e$ and $S_f$ is the user-selected strain energy fraction.  
  \begin{figure}
	\begin{subfigure}{0.30\textwidth}
		\centering
		\includegraphics[scale=.5]{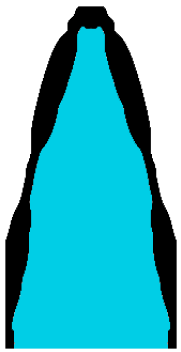}
		\caption{}\label{fig:2Dopt1}
	\end{subfigure}
	\begin{subfigure}{0.30\textwidth}
		\centering
		\includegraphics[scale=.5]{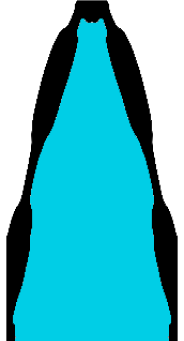}
		\caption{}\label{fig:2Dopt}
	\end{subfigure}
	\begin{subfigure}{0.30\textwidth}
		\centering
		\includegraphics[scale=0.28]{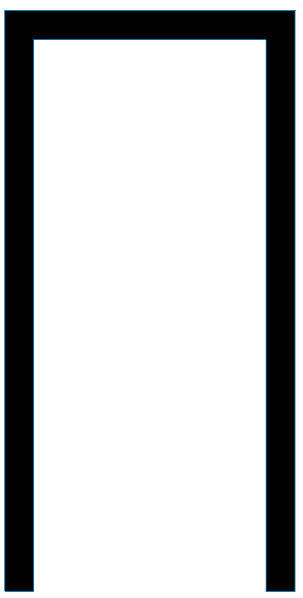}
		\caption{}
		\label{fig:2DrecCAD}
	\end{subfigure}
	\caption{\PKadd{2D optimized and rectangular chambers.  (\subref{fig:2Dopt1}) Optimized chamber I (\subref{fig:2Dopt}) Optimized chamber II, and (\subref{fig:2DrecCAD}) Rectangular chamber.}} \label{fig:Opt2DRect}
\end{figure} 

\begin{figure}
	\centering
	\begin{tikzpicture}
		\pgfplotsset{compat = 1.3}
		\begin{axis}[ blue,
			width = 0.7\textwidth,
			xlabel=MMA iteration,
			axis y line* = left,
			ylabel= Objective,
			ymajorgrids=true,
			xmajorgrids=true,
			grid style=dashed,
			legend style={at={(0.85,0.5)},anchor=east} ]
			\pgfplotstableread{Objb.txt}\mydata;
			\addplot[black, mark = square,mark size=1pt,style={very thick}]
			table {\mydata};
			\addlegendentry{Blueprint objective}
			\label{plot1}
		\end{axis}
		\begin{axis}[
			width = 0.7\textwidth,
			axis y line* = right,
			ylabel= Volume fraction,
			axis x line*= top,
			xlabel=MMA iteration,
			ytick = {0.025,0.075,0.125,0.15,0.175, 0.20, 0.225},
			yticklabel style={/pgf/number format/.cd,fixed,precision=3},
			legend style={at={(0.80,0.35)},anchor=east}]
			\pgfplotstableread{Volb.txt}\mydata;
			\addplot[smooth,red,mark = o,mark size=1pt,style={thick}]
			table {\mydata};
			\addlegendentry{Blueprint volume}
		\end{axis}
	\end{tikzpicture} 
	\caption{Objective and volume fraction convergence plots for the blueprint design \PKadd{of the optimized chamber II}}
	\label{fig:Convergenceplot}
\end{figure}
\subsection{Sensitivity analysis}
We use the Method of Moving Asymptotes (MMA)~\cite{svanberg1987method}, a gradient-based optimizer, for solving the formulated optimization problem (Eq.~\ref{eq:Optimizationequation}). Thus, derivatives of the objective and constraints with respect to the design variables are necessary; these are determined here using the adjoint-variable method.
\paragraph{Objective sensitivity}
The augmented performance function for the objective is written as

\begin{equation}\label{eq:augmentedperformance}
	\mathcal{L}_0 = f_\text{unit} + \trr{{^u}\bm{\lambda}}_q ( \mathbf{A}_q\mathbf{p}_q) +  \trr{{^p}\bm{\lambda}}_q \left(\mathbf{K}_q\mathbf{u}_q +\mathbf{T} \mathbf{p}_q\right),
\end{equation}
Differentiating Eq.~\ref{eq:augmentedperformance} with respect to the physical design variable $\bar{\rho_i}$, we get
\begin{equation}\label{eq:Lagrangeder}
	\begin{split}
		\frac{d \mathcal{L}_0}{d\bar{\rho}_i} = &\frac{\partial f_\text{unit}}{\partial \bar{\rho}_i} + 
		\frac{\partial f_{\text{unit}}}{\partial\mathbf{u}_q}\frac{\partial\mathbf{u}_q}{\partial \bar{\rho}_i}+ {^u}\bm{\lambda}_q^\top \left(\frac{\partial\mathbf{A}_q}{\partial\bar{\rho}_i}\mathbf{p}_q\right) + {^u}\bm{\lambda}_q^\top\left(\mathbf{A}_q\frac{\partial\mathbf{p}_q}{\partial\bar{\rho}_i}\right)\\ &+ {^p}\bm{\lambda}_q^\top \left(\frac{\partial\mathbf{K}_q}{\partial \bar{\rho}_i}\mathbf{u}_q + \mathbf{K}_q\frac{\partial\mathbf{u}_q}{\partial \bar{\rho}_i}\right) + {^p}\bm{\lambda}_q^\top \left(\frac{\partial\mathbf{T}}{\partial \bar{\rho}_i}\mathbf{p}_q + \mathbf{T}\frac{\partial\mathbf{p}_q}{\partial \bar{\rho}_i}\right)\\
		=&  \frac{\partial f_\text{unit}}{\partial \bar{\rho}_i}+{^p}\bm{\lambda}_q^\top \left(\frac{\partial\mathbf{K}_q}{\partial \bar{\rho}_i}\mathbf{u}_q\right) + {^u}\bm{\lambda}_q^\top \left(\frac{\partial\mathbf{A}_q}{\partial\bar{\rho}_i}\mathbf{p}_q\right) \\ &+ \underbrace{\left(\bm{l}^\top + {^p}\bm{\lambda}_q^\top \mathbf{K}_q\right)}_{\mathcal{T}_1}\frac{\partial\mathbf{u}_q}{\partial \bar{\rho}_i} + \underbrace{\left({^u}\bm{\lambda}_q^\top\mathbf{A}_q + {^p}\bm{\lambda}_q^\top \mathbf{T} \right)}_{\mathcal{T}_2}\frac{\partial\mathbf{p}_q}{\partial\bar{\rho}_i}.
	\end{split}
\end{equation} 
In context of adjoint-variable method, we choose ${^u}\bm{\lambda}_q$ and ${^p}\bm{\lambda}_q$ such that $\mathcal{T}_1=0$ and $\mathcal{T}_2=0$,  which give
\begin{equation} \label{eq:LagrangeMultiplier}
	\begin{aligned}
		{^p}\bm{\lambda}_q^\top = -\bm{l}^\top\inv{\mathbf{K}}_q,\quad
		{^u}\bm{\lambda}_q^\top = -{^p}\bm{\lambda}_q^\top \mathbf{T} \inv{\mathbf{A}}_q = \bm{l}^\top\inv{\mathbf{K}}_q\mathbf{T} \inv{\mathbf{A}}_q.
	\end{aligned}
\end{equation}
Using ${^u}\bm{\lambda}_q^\top $ and ${^p}\bm{\lambda}_q^\top $ in Eq.~\ref{eq:Lagrangeder} with $\frac{\partial f_\text{unit}}{\partial \bar{\rho}_i}=0$, we get
\begin{equation}\label{eq:senstivities_Obj}
	\frac{d f_\text{unit}}{d \bar{\rho_i}} = -\bm{l}^\top\inv{\mathbf{K}}_q \frac{\partial\mathbf{K}_q}{\partial \tilde{\rho}_i}\mathbf{u}_q + \underbrace{\bm{l}^\top\inv{\mathbf{K}}_q\mathbf{T} \inv{\mathbf{A}}_q\frac{\partial\mathbf{A}_q}{\partial\bar{\rho}_i}\mathbf{p}_q}_{\text{Pneumatic load sensitivities}}.
\end{equation}
One notes that due to the design-dependent nature of the pneumatic loads, $\frac{d f_\text{unit}}{d \bar{\rho_i}}$ contains load sensitivities (Eq.~\ref{eq:senstivities_Obj}). Thus, it should be included within the TO setting~\cite{kumar2020topology} while dealing with pneumatic actuation. Likewise, the derivatives of constraint g$_2$ with respect to the physical design variable $\bar{\rho_i}$ is evaluated as
\begin{equation}\label{Eq:senstivities_cons2}
	\frac{\partial {\text{g}_2}}{\partial \bar{\rho}_i} = \frac{-\frac{1}{2}\mathbf{u}^\top_q \frac{\partial\mathbf{K}_q}{\partial \bar{\rho}_i}\mathbf{u}_q + \mathbf{u}^\top_q \mathbf{T} \mathbf{A}^{-1}_q\frac{\partial\mathbf{A}_q}{\partial\bar{\rho}_i}\mathbf{p}_q}{se^*},
\end{equation}
whereas determining the derivative of g$_1$ is straightforward~\cite{kumar2022honeytop90}. To determine the derivative of the objective/constraints with respect to the design variables, the chain rule is applied as
\begin{equation}\label{Eq:objderivative}
	\frac{\partial{f}}{\partial \rho}_i = \frac{\partial {f}}{\partial \bar{\rho}_i}\frac{\partial \bar{\rho}_i}{\partial \tilde{\rho}_i} \frac{\partial \tilde{\rho}_i}{\partial {\rho}_i},
\end{equation}
where $f$ represents objective/constraint functions, $\frac{\partial \bar{\rho}_i}{\partial \tilde{\rho}_i}$ and $\frac{\partial \tilde{\rho}_i}{\partial {\rho}_i}$ are taken from Eq.~\ref{eq:derHeavy} and Eq.~\ref{eq:der_filter}, respectively. 

\begin{figure}[h!]
	\centering
	\begin{subfigure}[t]{0.15\textwidth}
		\centering
		\includegraphics[scale=0.5]{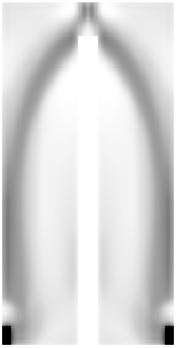}
		\caption{IN =5, $M_{nd} =44.65\%$, Obj = -160.53}
		\label{fig:IN5}
	\end{subfigure}
	\begin{subfigure}[t]{0.15\textwidth}
		\centering
		\includegraphics[scale=0.5]{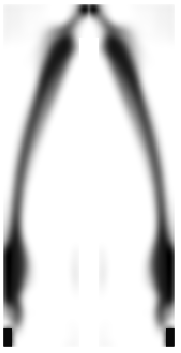}
		\caption{IN =10, $M_{nd} =23.82\%$, Obj = -391.84}
		\label{fig:IN10}
	\end{subfigure}
	\begin{subfigure}[t]{0.15\textwidth}
		\centering
		\includegraphics[scale=0.5]{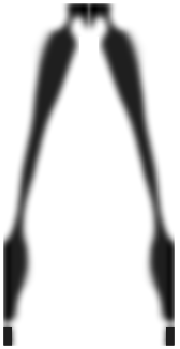}
		\caption{IN =20, $M_{nd} =15.11\%$, Obj = -684.99}
		\label{fig:IN20}
	\end{subfigure}
	\begin{subfigure}[t]{0.15\textwidth}
		\centering
		\includegraphics[scale=0.5]{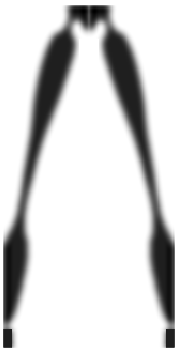}
		\caption{IN =30, $M_{nd} =14.20\%$, Obj = -717.41}
		\label{fig:IN30}
	\end{subfigure}
	\begin{subfigure}[t]{0.15\textwidth}
		\centering
		\includegraphics[scale=0.5]{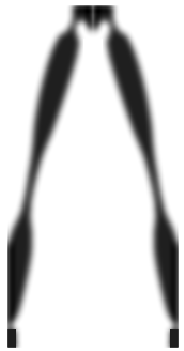}
		\caption{IN =50, $M_{nd} =14.01\%$, Obj = -728.47}
		\label{fig:IN50}
	\end{subfigure}
	\begin{subfigure}[t]{0.15\textwidth}
		\centering
		\includegraphics[scale=0.5]{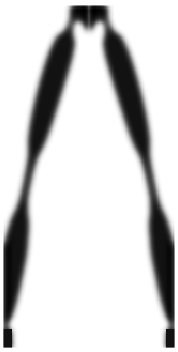}
		\caption{IN =100, $M_{nd} =12.56\%$, Obj = -727.24}
		\label{fig:IN100}
	\end{subfigure}
	\begin{subfigure}[t]{0.15\textwidth}
		\centering
		\includegraphics[scale=0.5]{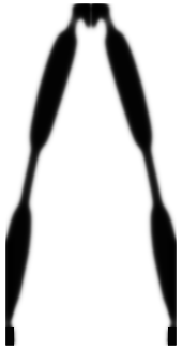}
		\caption{IN =150, $M_{nd} =9.01\%$, Obj = -728.67}
		\label{fig:IN150}
	\end{subfigure}
	\begin{subfigure}[t]{0.15\textwidth}
		\centering
		\includegraphics[scale=0.5]{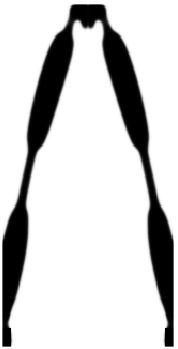}
		\caption{IN =200, $M_{nd} =4.74\%$, Obj = -740.37}
		\label{fig:IN200}
	\end{subfigure}
	\begin{subfigure}[t]{0.15\textwidth}
		\centering
		\includegraphics[scale=0.5]{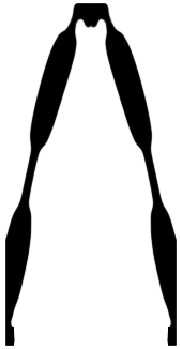}
		\caption{IN =250, $M_{nd} =2.26\%$, Obj = -723.24}
		\label{fig:IN250}
	\end{subfigure}
	\begin{subfigure}[t]{0.15\textwidth}
		\centering
		\includegraphics[scale=0.5]{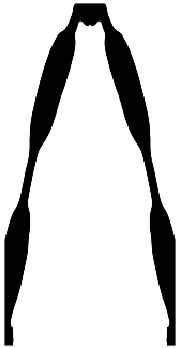}
		\caption{IN =300, $M_{nd} =1.11\%$, Obj = -7612.53}
		\label{fig:IN300}
	\end{subfigure}
	\begin{subfigure}[t]{0.15\textwidth}
		\centering
		\includegraphics[scale=0.5]{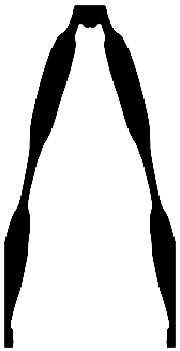}
		\caption{IN =350, $M_{nd} =0.55\%$, Obj = -598.48}
		\label{fig:IN350}
	\end{subfigure}
	\begin{subfigure}[t]{0.15\textwidth}
		\centering
		\includegraphics[scale=0.5]{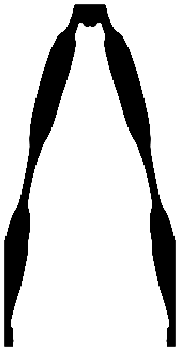}
		\caption{IN =373, $M_{nd} =0.28\%$, Obj = -598.83}
		\label{fig:IN373}
	\end{subfigure}
	\caption{Design evolution at different MMA iterations for optimized chamber II. IN, $M_{nd}$, and Obj indicate the iteration number, the discreteness measure, and the objective value, respectively.} \label{fig:Optresultsiterations}
\end{figure} 

\begin{figure}
	\begin{subfigure}{0.30\textwidth}
		\centering
		\includegraphics[scale=0.23]{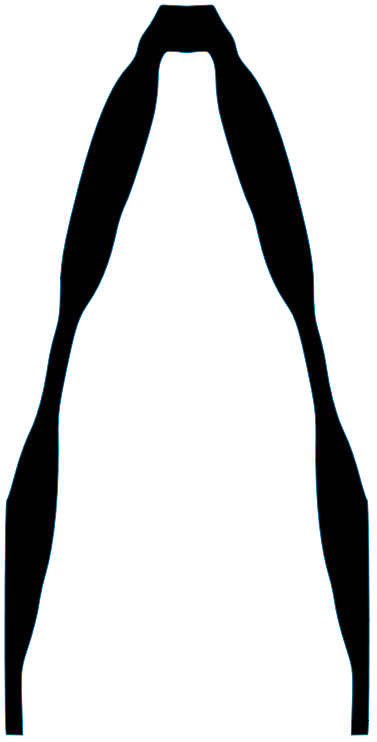}
		\caption{}
		\label{fig:2D70optCAD}
	\end{subfigure}
	\begin{subfigure}{0.30\textwidth}
		\centering
		\includegraphics[scale=0.23]{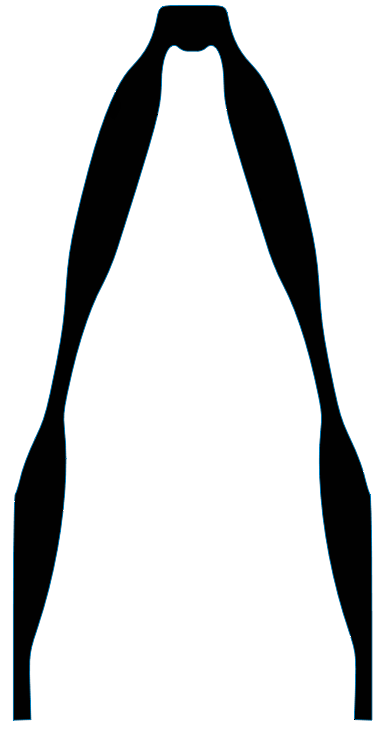}
		\caption{}
		\label{fig:2DoptCAD}
	\end{subfigure}
	\begin{subfigure}{0.30\textwidth}
		\centering
		\includegraphics[scale=0.28]{rec2DCAD}
		\caption{}
		\label{fig:2DrecCAD1}
	\end{subfigure}
	\caption{\PKadd{CAD models are displayed.  (\subref{fig:2Dopt1}) Optimized chamber~I (\subref{fig:2Dopt}) Optimized chamber~II and (\subref{fig:2DrecCAD1}) Rectangular chamber.}} \label{fig:Opt2DRectCAD}
\end{figure} 
\begin{figure}
	\centering
	\begin{subfigure}[t]{0.3\textwidth}
		\centering
		\includegraphics[scale=0.7]{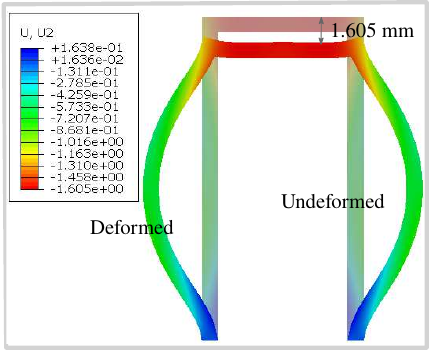}
		\caption{Conventional chamber}
		\label{fig:2D_rec_result}
	\end{subfigure}
	\quad
	\begin{subfigure}[t]{0.3\textwidth}
		\centering
		\includegraphics[scale=0.65]{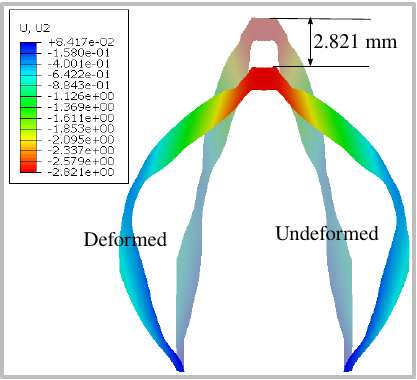}
		\caption{Optimized chamber 1}
		\label{fig:2D_opt_result70}
	\end{subfigure}
	\quad
	\begin{subfigure}[t]{0.3\textwidth}
		\centering
		\includegraphics[scale=0.65]{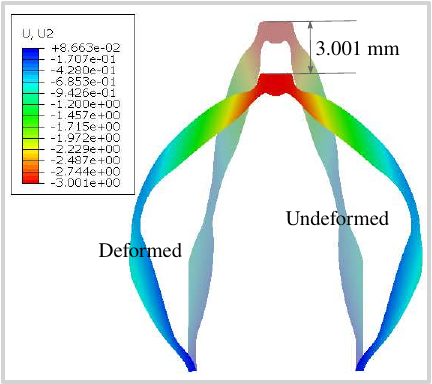}
		\caption{Optimized chamber 2}
		\label{fig:2D_opt_result}
	\end{subfigure}
	\caption{Performance comparison of conventional rectangular and optimized chambers. (\subref{fig:2D_rec_result}), (\subref{fig:2D_opt_result70}) and (\subref{fig:2D_opt_result}) depict the undeformed and deformed profiles for the conventional chamber,  optimized chamber I, and optimized chamber II, respectively. The finite element analyses are performed in ABAQUS with geometric and material nonlinearity. The optimized chamber II provides 86.98\% more deformation than the conventional chamber at a $\SI{10}{\kilo\pascal}$ pressure load. U2 indicates deformation in the $y-$direction.} \label{fig:2DAnalysis}
\end{figure} 

\begin{figure}
	\centering
	\begin{subfigure}[t]{0.15\textwidth}
		\centering
		\includegraphics[scale=0.65]{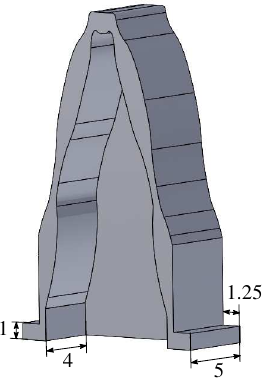}
		\caption{Iso view}
		\label{fig:unit_iso}
	\end{subfigure}
	\quad
	\begin{subfigure}[t]{0.15\textwidth}
		\centering
		\includegraphics[scale=0.62]{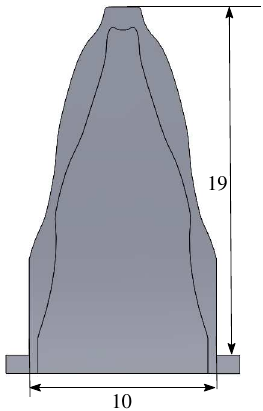}
		\caption{Front view}
		\label{fig:unit_front}
	\end{subfigure}
	\begin{subfigure}[t]{0.15\textwidth}
		\centering
		\includegraphics[scale=0.25]{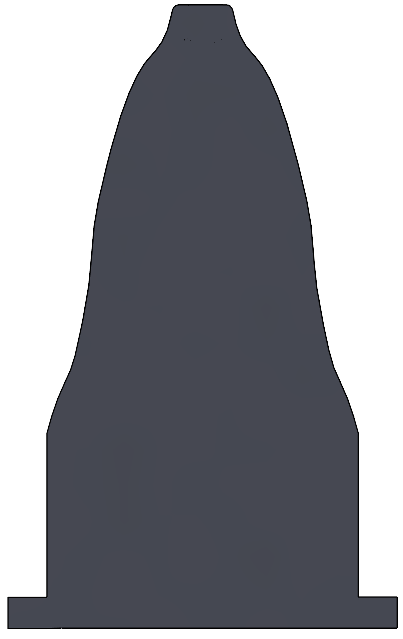}
		\caption{Back view}
		\label{fig:unit_back}
	\end{subfigure}
	\caption{Half sectional view of a 3D unit pressure chamber with dimensions in mm.} \label{fig:3Dunit_dim}
\end{figure} 
\begin{figure}
	\centering
	\includegraphics[scale=1.25]{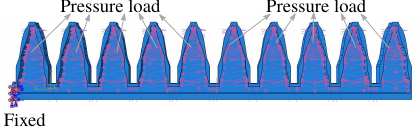}
	\caption{A section view of an arm of the optimized PSG is depicted with the applied pressure load and fixed boundary conditions. A symmetry boundary condition is applied in the normal direction to the section.}\label{fig:3Dbound}
\end{figure}

  
 \section{Optimized PneuNet design}\label{Sec:OptimizedUnit}

The SHDD for a PneuNet is depicted in Fig.~\ref{fig:DD} with output location ``O ." The pneumatic load and displacement boundary conditions are also depicted. $L_x$ and $L_y$ indicate the dimension in $x-$ and $y-$directions, respectively. The figure also indicates the non-design solid and void regions, that do not change as optimization progresses.

The reference symmetric domain is parameterized using $\texttt{nelx}\times \texttt{nely} = 80 \times 320$ bilinear FEs. The volume fraction of the blueprint design is set to 0.20. We consider two cases: Case~I and Case~II, while imposing the strain energy constraint on the corresponding eroded design. $S_{f}$ for Case I and Case II are set to $0.70$ and $0.90$, respectively. The SIMP parameter $n=3$ is taken. The filter radius $r_\textit{fill}$ =7.6 is set. The workpiece stiffness $k_s = 1$  is set at the output location. The Darcy and drainage parameters $\left\{\eta_k,\,\beta_k\right\}$ = $\left\{\eta_d,\,\beta_d\right\}$ = $\left\{0.1,\,10\right\}$ are taken. The maximum value of the projection parameter, $\beta_\textit{max}=128$, is set, and it is doubled every 50 MMA iterations. The maximum number of MMA iterations is set to 400. The external move limit for the MMA is set to $0.1$. The given volume fraction is used as the initial guess for the optimization process. The optimization continues until the maximum number of iterations is reached or the maximum absolute change in the design vectors between two successive iterations comes below 0.0001.

The optimized pressure chamber I (Case I) and optimized pressure chamber II (Case II) are shown in Fig.~\ref{fig:2Dopt1} and Fig.~\ref{fig:2Dopt}, respectively. The former and later chambers converge at the 366th and 373rd MMA iterations, respectively. The optimized chambers are different in shape than a conventional rectangular-shaped pressure chamber depicted in Fig.~\ref{fig:2DrecCAD}. One notes that the optimized designs are free from hinge/single-node connections. Although the optimized units (Fig.~\ref{fig:2Dopt1} and Fig.~\ref{fig:2Dopt}) appear as single connected chambers, these are  results of the TO formulation provided in Eq.~\ref{eq:Optimizationequation}, with no prior assumptions on the number of sub-chambers~(holes), their connectivity, or material layout. Unlike shape optimization, which requires a predefined topology and thus restricts the design search space, TO permits free material evolution within the domain and can yield superior configurations. Additionally, the proposed method is general and extendable to multi-material and/or multifunctional soft pneumatic grippers or other related applications, where topology change cannot be handled efficiently by shape optimization. 

The convergence plots for objective and volume constraint for optimized chamber II are shown in Fig.~\ref{fig:Convergenceplot}. One notes that the volume constraint remains active at the end of the optimization. Jumps in the convergence plots are due to the $\beta$-updating scheme, as expected: an increase in $\beta$ changes the gradients of the objective and constraints.

Figure~\ref{fig:Optresultsiterations} indicates designs,  $M_{nd}$ (discreteness measure), and objective values at different MMA iterations for the optimized chamber II. The objective values and  $M_{nd}$ for the designs at 200th and 373rd iterations are $\{-740.37,\,4.74\%\}$ and $\{ -598.83,\,0.28\%\}$, respectively (Fig.~\ref{fig:Optresultsiterations}). Although the objective value at the 200th iteration is higher than that of the 373rd iteration, the corresponding $M_{nd}$ is significantly larger, indicating that many elements still possess intermediate physical density $0<\bar{\rho}<1$ and remain in a transitional state of the optimization. Therefore, the design at the 200th iteration lacks a clear physical interpretation and is not suitable for CAD reconstruction (Fig.~\ref{fig:Optresultsiterations}). However, the design at the final iteration exhibits a significantly lower $M_{nd}$, indicating a near 0-1 solution. This improved discreteness facilitates the generation of a suitable CAD model and subsequent fabrication. Thus, the final optimized design is used for further study.

\subsection{2D CAD models and FEM analyses}
As we employ the robust formulation, the obtained optimized design is close to 0-1. The steps mentioned in~\cite{kumar2022topological} are used to create the \texttt{DXF} file, which is imported into the \texttt{SolidWorks} software to create the 2D CAD model. For ABAQUS 2D nonlinear FE analyses, the 2D CAD models (Fig.~\ref{fig:Opt2DRectCAD}) are exported as  \texttt{STEP} files. 
\subsection{2D Performance comparison}
This section provides a comparative study of the performances of the optimized and conventional chambers. These chambers are subjected to the same internal pressure/pneumatic loads and are modeled with identical boundary conditions in ABAQUS. Both geometric and material nonlinearities with plane-stress conditions are considered. 
The 3D arms of SPG are printed using  \textit{Elastic 50A Resin V2}, a soft, rubber-like elastic. Thus, the Ogden hyperelastic material model that captures highly nonlinear stress-strain relations and is well-suited for rubber-like elastic material,  is employed~\cite{yao2022generalized}. The strain energy function $W$ of the material is defined as~\cite{abaqus2024}
\begin{equation}\label{Eq:OgdenMM}
	W = \sum_{i=1}^{N} \frac{2\mu_i}{\alpha_i^2} \left( \bar{\lambda}_1^{\alpha_i} + \bar{\lambda}_2^{\alpha_i} + \bar{\lambda}_3^{\alpha_i} - 3 \right) + \sum_{i=1}^{N} \frac{1}{D_i} \left( J - 1 \right)^{2i}
\end{equation}
where $\lambda_i|_{i = 1,\,2,\,3}$ are the principal stretches. $\mu_i$, $\alpha_i$, and $D_i$ are the material parameters. The first and latter parameters are related to the material's shear and bulk modulus, respectively; whereas $\alpha_i$ controls the nonlinearity of the stress-strain curve~\cite{abaqus2024}. $N$ indicates the order of the Ogden model. Order 1, i.e., $N=1$  material model is used with $\mu_1 = 1.57$, $\alpha_1 =2$ and $D_1 = 5\times 10^{-6}$. These parameters are determined using a material calibration process. The parameters are calibrated against experimental stress-strain data provided by Formlabs~\cite{FormlabsElastic50AResinV2}. The constants $\mu_1, \alpha_1, D_1$ are obtained by fitting the Ogden constitutive relation per ABAQUS~\cite{abaqus2024} to the experimental curves by minimizing the least-square error, $||\sigma_\text{Ogden}(\epsilon, \mu_1, \alpha_1, D_1) -\sigma_\text{Exp} (\epsilon)||^2$ using the Levenberg-Marquardt approach~\cite{marechal2021toward}.

We apply pneumatic pressure of  \SI{10} {\kilo \pascal} within the optimized and conventional chambers to perform nonlinear finite element analyses in ABAQUS, while their bottom edges are fixed. The results of the studies, including deformed profiles and their undeformed counterparts, are depicted in Fig.~\ref{fig:2DAnalysis}. The optimized chamber I, optimized chamber II and rectangular chamber provide $y-$direction deformations of \SI{2.821}{\milli\meter}, \SI{3.001}{\milli\meter} and \SI{1.605}{\milli\meter}, respectively, indicating that the optimized chamber I and optimized chamber II  yield approximately 75.76\%, and 86.98\% greater deformation than their conventional counterpart. This means that a gripper with optimized chambers can provide greater gripping force than a traditional design. Additionally, as the optimized chamber~II provides relatively more deformation than the optimized chamber I, we use the former for 3D analyses, fabrication, and experimental demonstration. The following sections present 3D analysis results, the 3D printing workflow, experimental setup, and experimental grasping examples.

\begin{table*}
	\caption{\PKadd{The deformation profiles of the arms with both optimized and conventional chambers are shown at different pneumatic loads, with and without gravity. The corresponding tip displacements in the $ x$- and $ y$-directions ($u_x$ and $u_y$) are reported for both configurations. In the deformed profiles, the color contours represent deformation magnitude, where blue and red denote the minimum and maximum deformation regions, respectively.}}\label{Tab:3DdeformationProfiles}
	\resizebox{\textwidth}{!}{%
	\begin{tabular}{|cc|c|c|c|c|c|c|}
		\hline
		\multicolumn{2}{|c|}{\begin{tabular}[c]{@{}c@{}}\textbf{Pressure load}\\ (kPa)\end{tabular}}                                                                                                     & $\mathbf{60}$      & $\mathbf{90 }$     & $\mathbf{100 }$    & $\mathbf{125}$     & $\mathbf{150}$    & $\mathbf{175}$    \\ \hline
		\multicolumn{1}{|c|}{\multirow{6}{*}{\begin{tabular}[c]{@{}c@{}}\textbf{Without gravity}\\ \textbf{deformation profiles}\end{tabular}}} & \begin{tabular}[c]{@{}c@{}}\textbf{Optimized}\\ \textbf{arm}\end{tabular}         &     \rule{0pt}{1.5cm}\includegraphics[scale=0.08]{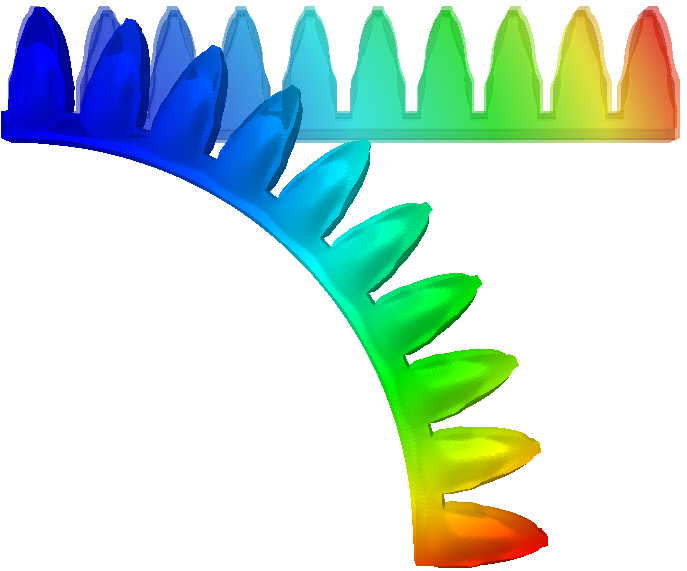} &  \rule{0pt}{1.5cm}\includegraphics[scale=0.08]{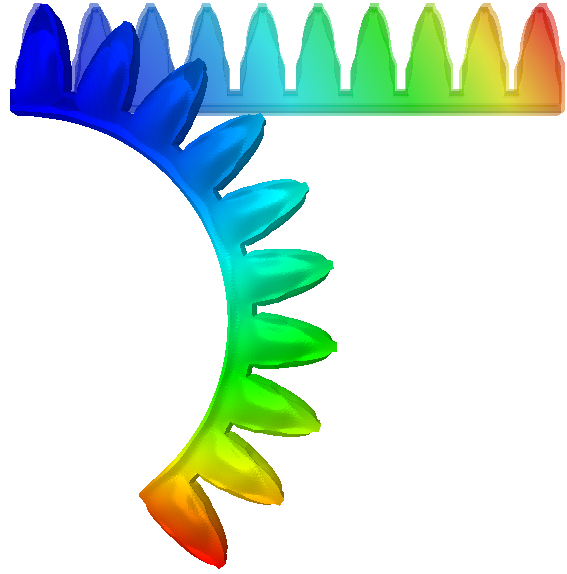} & \rule{0pt}{1.5cm}\includegraphics[scale=0.08]{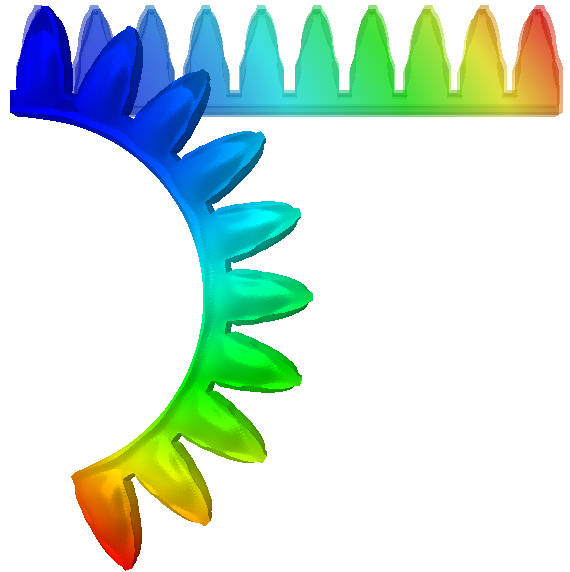} & \rule{0pt}{1.5cm}\includegraphics[scale=0.08]{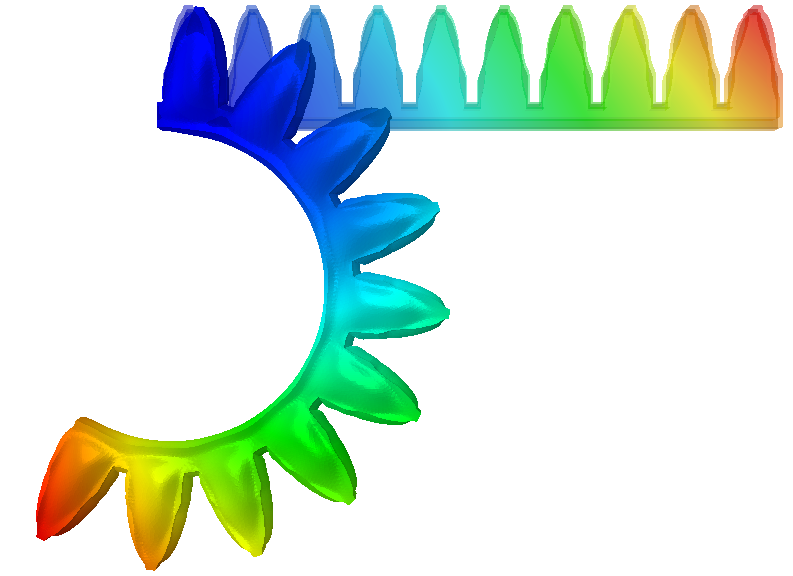} & \rule{0pt}{1.5cm}\includegraphics[scale=0.08]{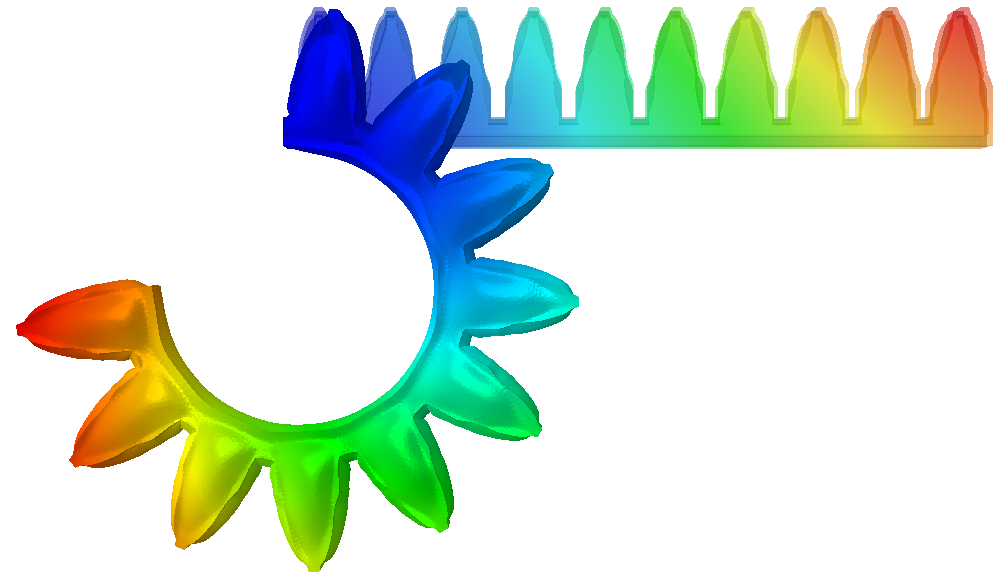} & \rule{0pt}{1.5cm}\includegraphics[scale=0.09]{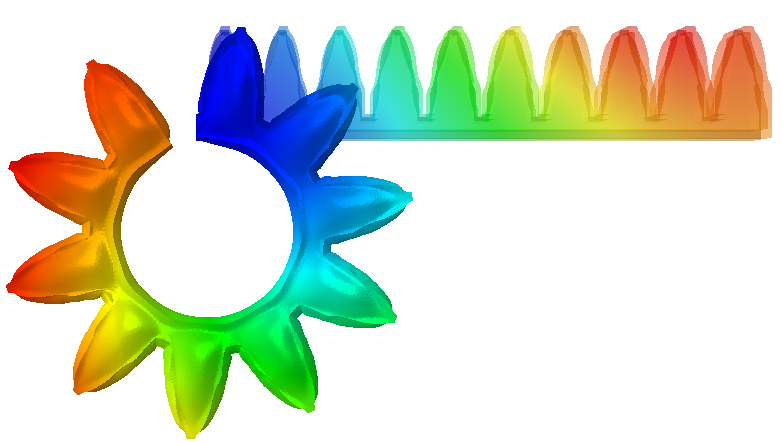}    \\ \cline{2-8} 
		\multicolumn{1}{|c|}{}                                                                                                & $u_x$ (mm)                                                         & 20.30   & 73.13   & 94.01   & 143.23  & 165.73 & 141.53 \\ \cline{2-8} 
		\multicolumn{1}{|c|}{}                                                                                                & $u_y$ (mm)                                                         & -97.94  & -125.48 & -126.12 & -104.39 & -54.98 & -7.10  \\ \cline{2-8} 
		\multicolumn{1}{|c|}{}                                                                                                & \begin{tabular}[c]{@{}c@{}}\textbf{Conventional}\\ \textbf{arm}\end{tabular}      &\rule{0pt}{1.5cm}\includegraphics[scale=0.08]{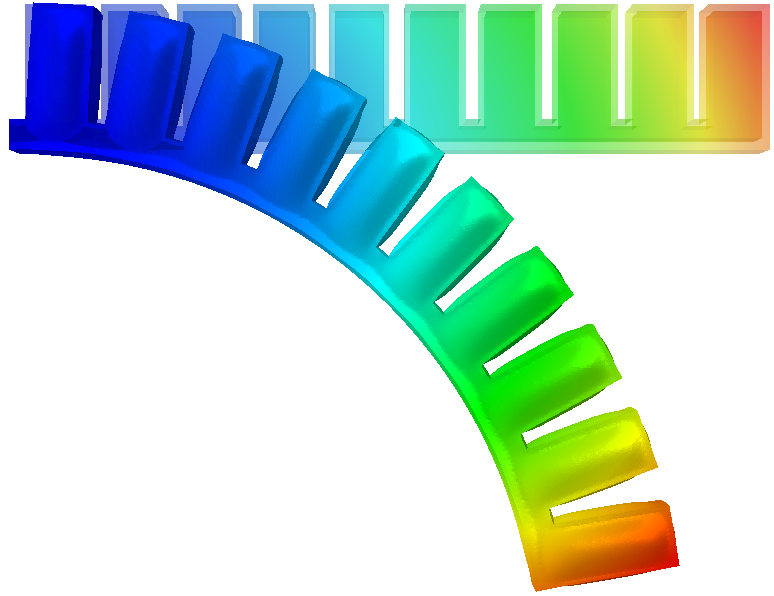} &  \rule{0pt}{1.5cm}\includegraphics[scale=0.08]{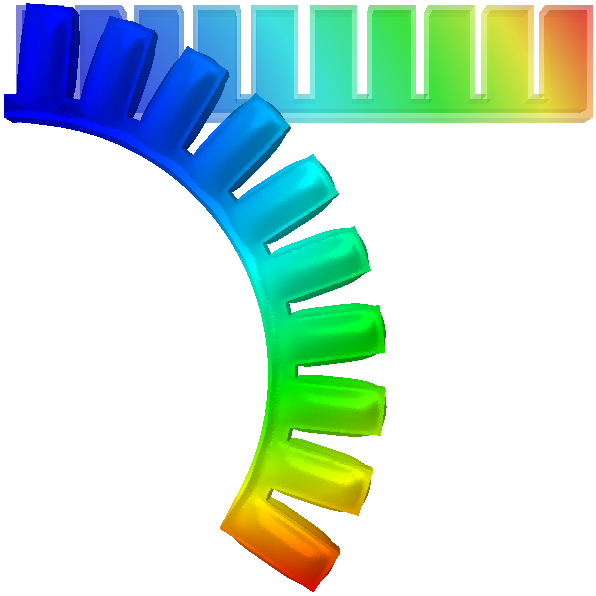} & \rule{0pt}{1.5cm}\includegraphics[scale=0.08]{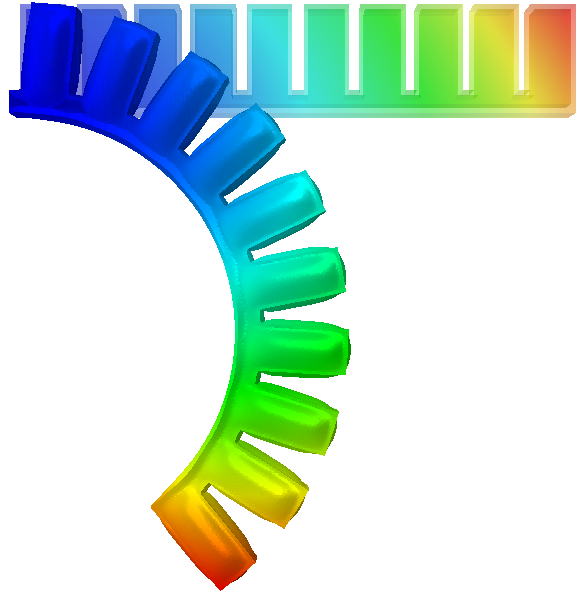} & \rule{0pt}{1.5cm}\includegraphics[scale=0.08]{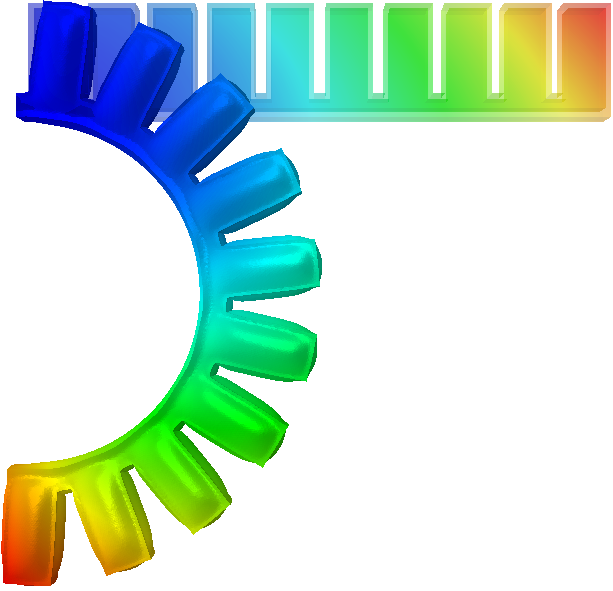} & \rule{0pt}{1.5cm}\includegraphics[scale=0.08]{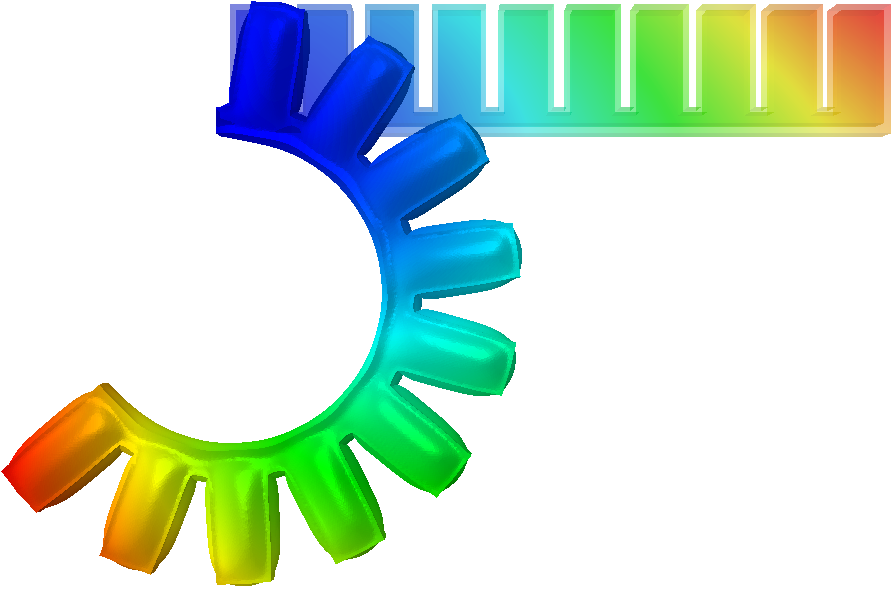} & \rule{0pt}{1.5cm}\includegraphics[scale=0.08]{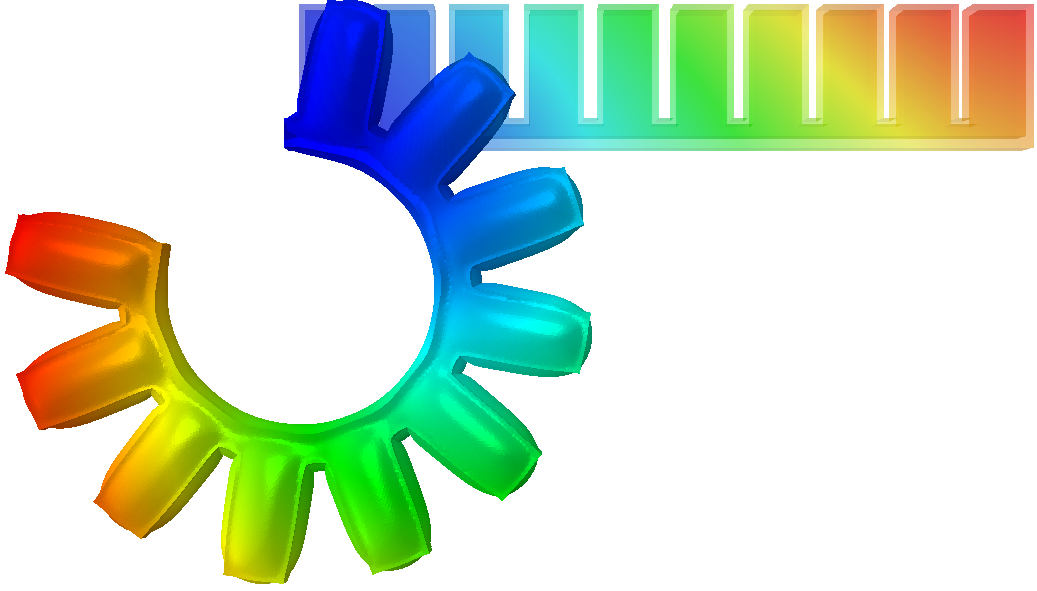}         \\ \cline{2-8} 
		\multicolumn{1}{|c|}{}                                                                                                & $u_x$ (mm)                                                         & 14.79   & 59.36   & 78.39   & 127.80  & 164.90 & 169.19 \\ \cline{2-8} 
		\multicolumn{1}{|c|}{}                                                                                                & $u_y$ (mm)                                                         & -92.44  & -124.88 & -129.45 & -122.0  & -86.60 & -35.15 \\ \hline
		\multicolumn{1}{|c|}{\multirow{6}{*}{\begin{tabular}[c]{@{}c@{}}\textbf{With gravity}\\ \textbf{deformation profiles}\end{tabular}}}    & \begin{tabular}[c]{@{}c@{}}\textbf{Optimized}\\ \textbf{arm}\end{tabular}        &  \rule{0pt}{1.5cm}\includegraphics[scale=0.08]{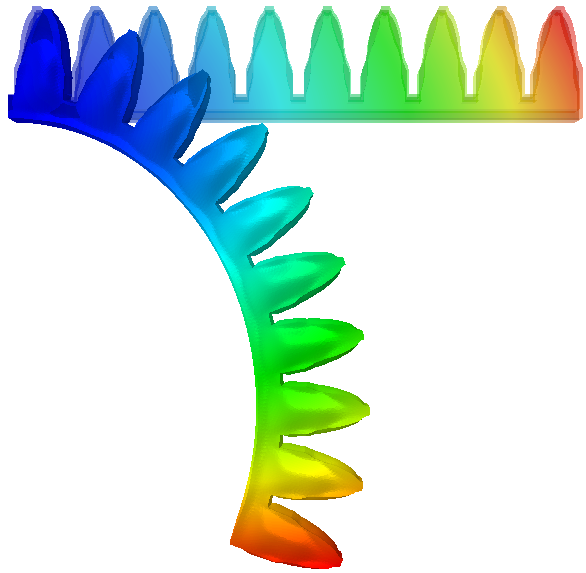} & \rule{0pt}{1.5cm}\includegraphics[scale=0.08]{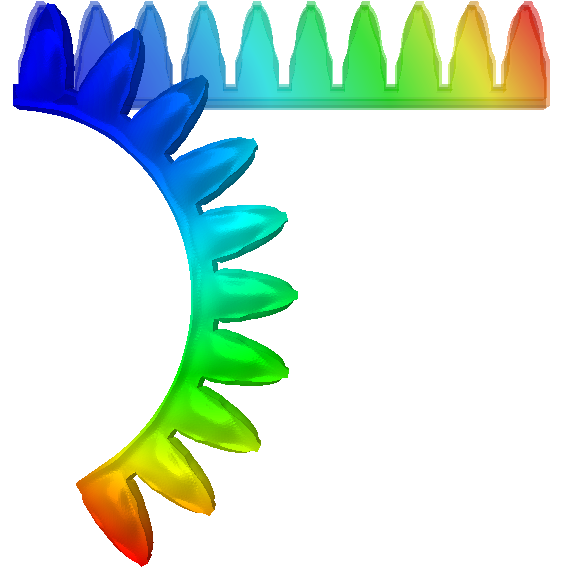} & \rule{0pt}{1.5cm}\includegraphics[scale=0.08]{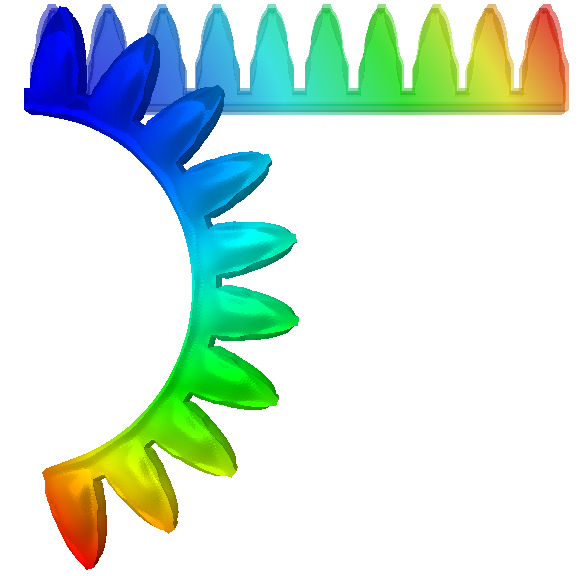} & \rule{0pt}{1.5cm}\includegraphics[scale=0.08]{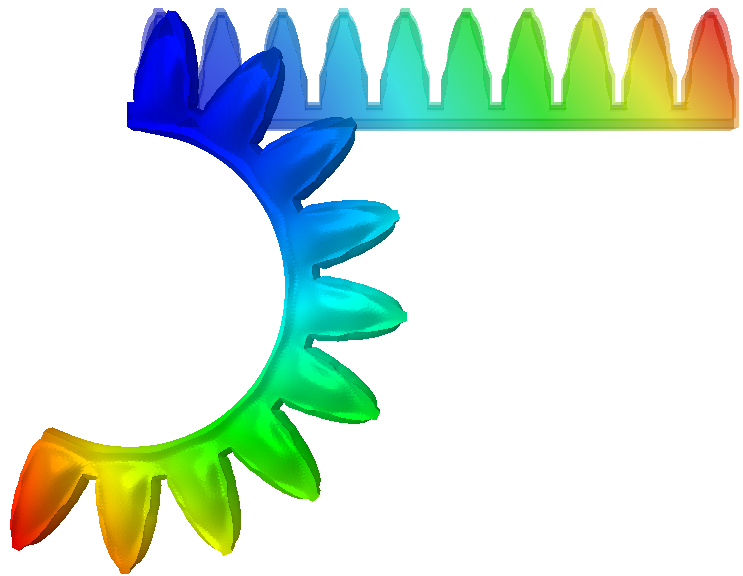} & \rule{0pt}{1.5cm}\includegraphics[scale=0.08]{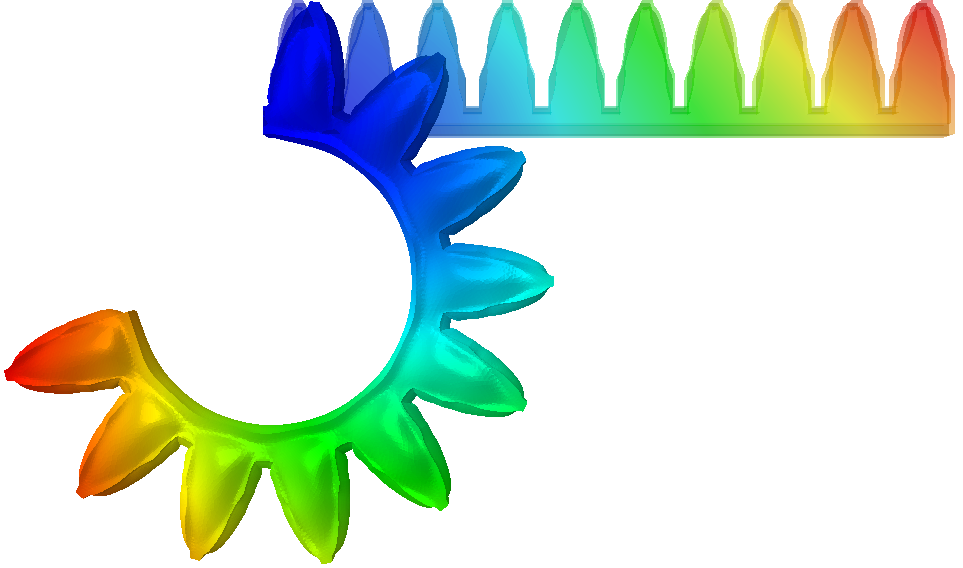} & \rule{0pt}{1.5cm}\includegraphics[scale=0.08]{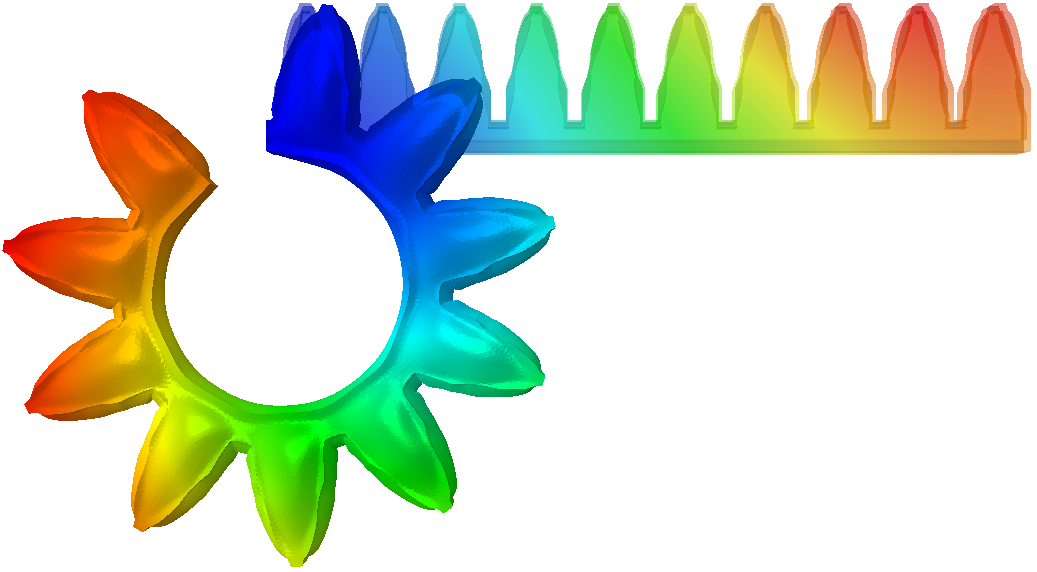}          \\ \cline{2-8} 
		\multicolumn{1}{|c|}{}                                                                                                & $u_x$ (mm)                                                         & 48.33   & 90.10   & 105.59  & 142.92  & 165.56 & 148.06 \\ \cline{2-8} 
		\multicolumn{1}{|c|}{}                                                                                                & $u_y$ (mm)                                                         & -120.67 & -130.22 & -128.34 & -108.45 & -65.97 & -13.90 \\ \cline{2-8} 
		\multicolumn{1}{|c|}{}                                                                                                & \begin{tabular}[c]{@{}c@{}}\textbf{Conventional}\\     \textbf{arm}\end{tabular} &  \rule{0pt}{1.5cm}\includegraphics[scale=0.08]{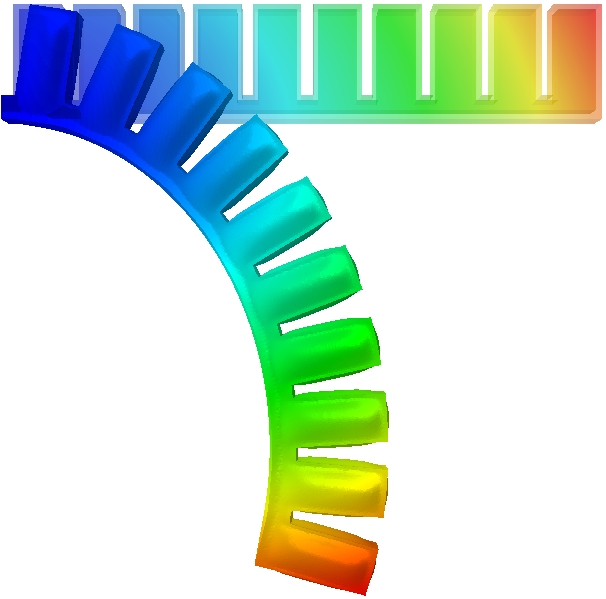} & \rule{0pt}{1.5cm}\includegraphics[scale=0.08]{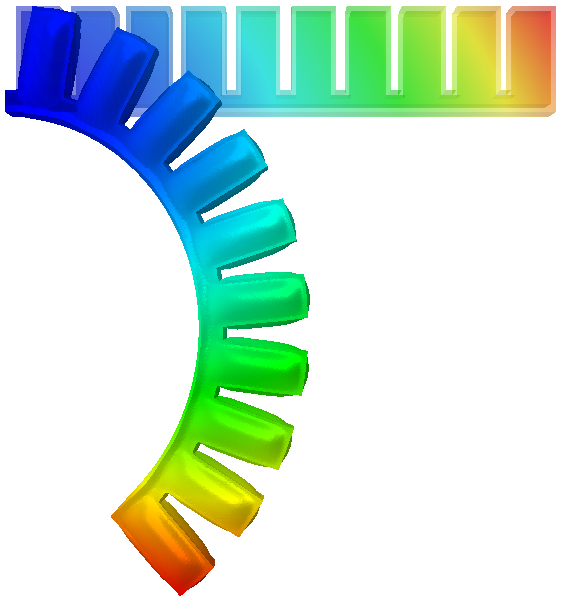} & \rule{0pt}{1.5cm}\includegraphics[scale=0.08]{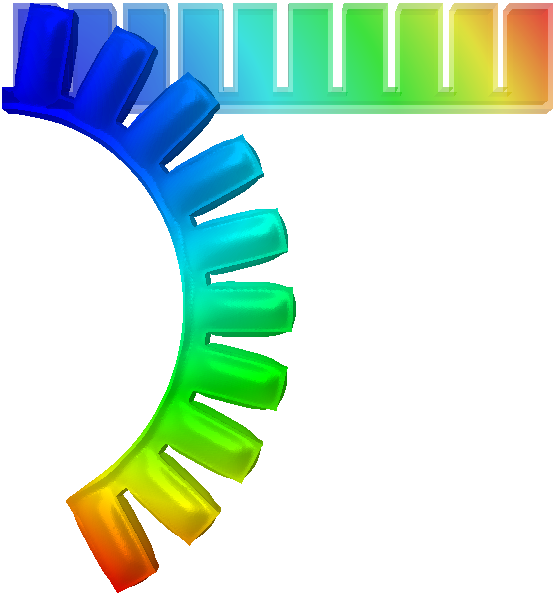} & \rule{0pt}{1.5cm}\includegraphics[scale=0.08]{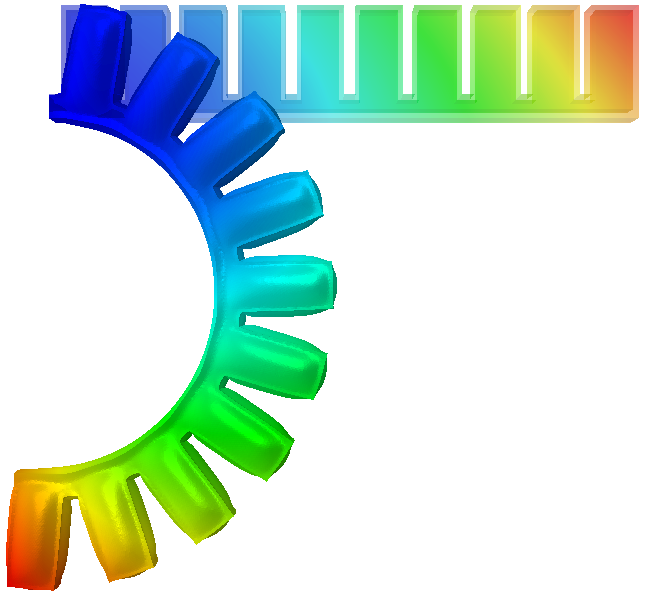} & \rule{0pt}{1.5cm}\includegraphics[scale=0.08]{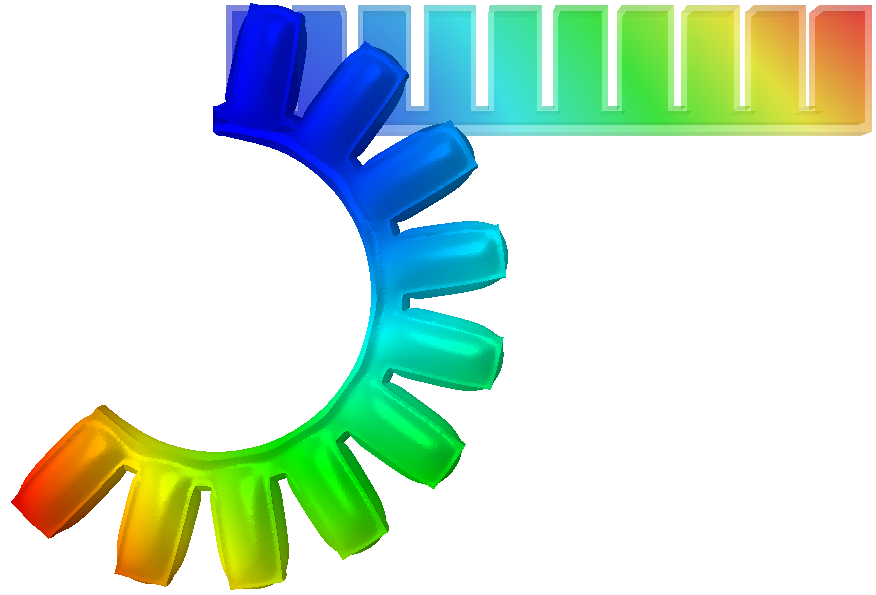} & \rule{0pt}{1cm}\includegraphics[scale=0.08]{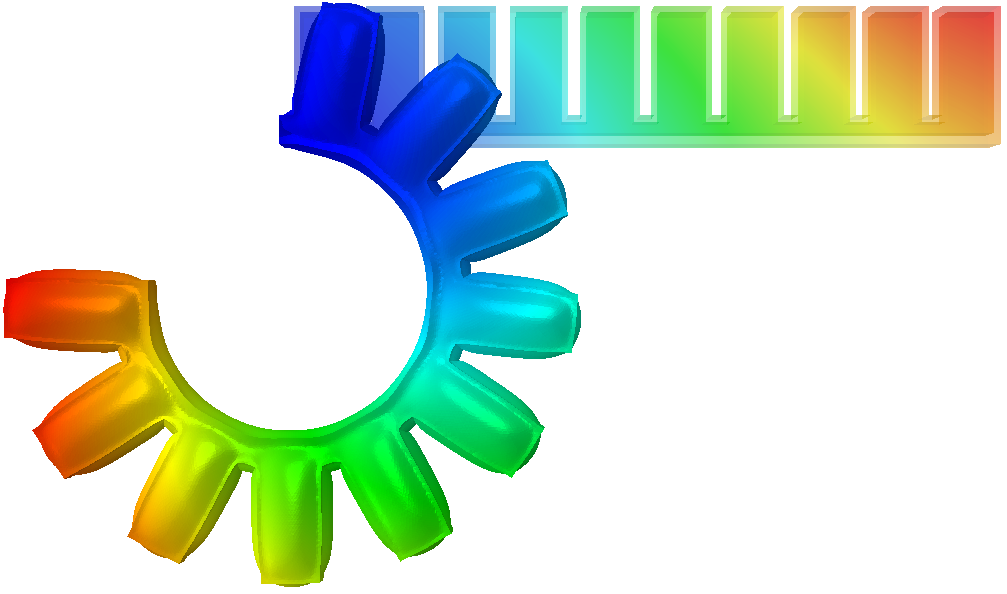}     \\ \cline{2-8} 
		\multicolumn{1}{|c|}{}                                                                                                & $u_x$ (mm)                                                         & 49.17   & 85.25   & 98.89   & 133.92  & 163.43 & 172.23 \\ \cline{2-8} 
		\multicolumn{1}{|c|}{}                                                                                                & $u_y$ (mm)                                                         & -123.21 & -133.96 & -133.93 & -123.26 & -94.26 & -47.35 \\ \hline
	\end{tabular}
}
\end{table*}

\section{3D  soft pneumatic grippers}\label{Sec:3DSPG}
The 2D CAD model of the optimized chamber is extruded 5 mm to create a 3D CAD model, which is depicted in Fig.~\ref{fig:3Dunit_dim} with all other dimensions~\cite{lu2022optimal}. Ten such units (Fig.~\ref{fig:unit_iso}) are assembled to form an arm for the proposed SPG as shown in Fig.~\ref{fig:3D_CAD_opt}. Note that the 3D arm requires less back-face area than conventional 3D design with rectangular chambers (Fig.~\ref{fig:unit_back}). This facilitates 3D printing by minimizing the need for support structures and thus, cost. 

\begin{figure*}
	\includegraphics[scale=2]{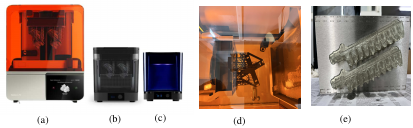}
	\caption{(a) Form 4 3D SLA printer (b) Form wash machine (c) Form cure machine (d)  Printing of support structures (Fig.~\ref{fig:SPG_four_def}) (e) Cured 3D-printed gripper arms. Form Wash and Form Cure are used to print, wash, and cure the 3D-printed arms.}\label{fig:3D-printer}
\end{figure*}

\subsection{FEM simulation}
This section provides a finite element framework used to determine the deformation behavior of the SPG with optimized chambers. The simulations are performed with different pressure loading in ABAQUS, incorporating large deformation effects, self-contact, and material incopressibility. Ogden material model (Eq.~\ref{Eq:OgdenMM}) is used. 

Fig.~\ref{fig:3Dbound} depicts a cross-sectional view with the fixed boundary conditions. The pressure load is applied on the internal surfaces representing the arm's inlet and chamber walls (Fig.~\ref{fig:3Dbound}). A symmetric boundary condition is applied normal to the sectional plane. Material density is set to $\SI{1.01e-6}{\kilogram\per\milli\meter\cubed}$. The gripper is assumed to be isotropic and nearly incompressible, which is consistent with the characteristics of the elastomers used for 3D printing. The pneumatic loading is modeled as a quasi-static process wherein the load is applied incrementally to mimic the gradual inflation. The gripper is discretized using 8-noded hybrid elements (C3D8RH) that account for the nearly incompressible behavior of the elastic material~\cite{abaqus2024}. C3D8RH FEs are specifically designed to reduce volumetric locking under large strains and have an independent pressure degree of freedom~\cite{abaqus2024}. 

Table~\ref{Tab:3DdeformationProfiles} demonstrates the deformation profiles for a soft pneumatic arm made via the optimized and conventional rectangular chambers at different pressure loads with and without gravity (self-weight). The tip $x$ and $y$ deformation values are shown. One notes the arm made of the optimized chambers exhibits greater bending deformation under the applied loads than the one made with the conventional chambers, i.e., the arm with optimized chambers outperforms one with the traditional chambers. Additionally, when the deformed profiles form an angle $90^\circ$ (clockwise) with the undeformed configurations, the arms exhibit greater deformation under the influence of gravity. However, when the angle exceeds $90^\circ$, the gravitational load begins to resist the upward deformation of the arm, resulting in reduced overall displacement.

Note that the extrusion of the optimized unit into a 3D configuration for constructing a 3D arm is based on the assumption of a constant cross-section. With this assumption,  the desired bending deformation is governed by the in-plane topology of the cross-section, as also indicated by the bending deformation profiles at different applied pressure loads shown in Table.~\ref{Tab:3DdeformationProfiles}. Therefore, the extrusion of a 2D-optimized unit into 3D preserves the intended bending mode shapes for the gripping action. Additionally, the front and back covers are introduced in the 3D model (Fig.~\ref{fig:3Dunit_dim}) purely to ensure airtightness and manufacturability. With a thickness of 1 mm (Fig.~\ref{fig:3Dunit_dim}), which is small relative to the chamber dimension ($10\times 10\times 20$ mm$^3$), their contribution to the bending stiffness is negligible and does not hinder the desired bending deformation as indicated by the deformation profiles at various pressure loads in Table~\ref{Tab:3DdeformationProfiles}.
\begin{table}
	\caption{\PKadd{Numerical and experimental deformation profiles of the 3D arm with different pneumatic loads are presented. The maximum horizontal tip displacement, $u_x$, is reported for both cases, and relative errors are evaluated to assess agreement between simulations and experiments. In the numerically obtained deformed profiles, the color contours indicate deformation magnitude, with blue and red denoting the minimum and maximum deformation regions, respectively.}}\label{Tab:3DdeformationNumExp}
		\resizebox{\textwidth}{!}{%
	\begin{tabular}{|c|cccc|c|}
		\hline
		\multirow{2}{*}{\begin{tabular}[c]{@{}c@{}}\textbf{Applied Pressure}\\ (kPa)\end{tabular}} & \multicolumn{4}{c|}{\textbf{Deformation Profiles}}                                                 & \multirow{2}{*}{\textbf{Error} (\%)} \\ \cline{2-5}
		& \multicolumn{1}{c|}{\textbf{Experiment}} & \multicolumn{1}{c|}{${u}_x$ (mm)} & \multicolumn{1}{c|}{\textbf{Numerical}} & ${u}_x$ (mm)&                             \\ [1em]\hline
		0                                                                                 & \multicolumn{1}{c|}{\includegraphics[scale=0.15]{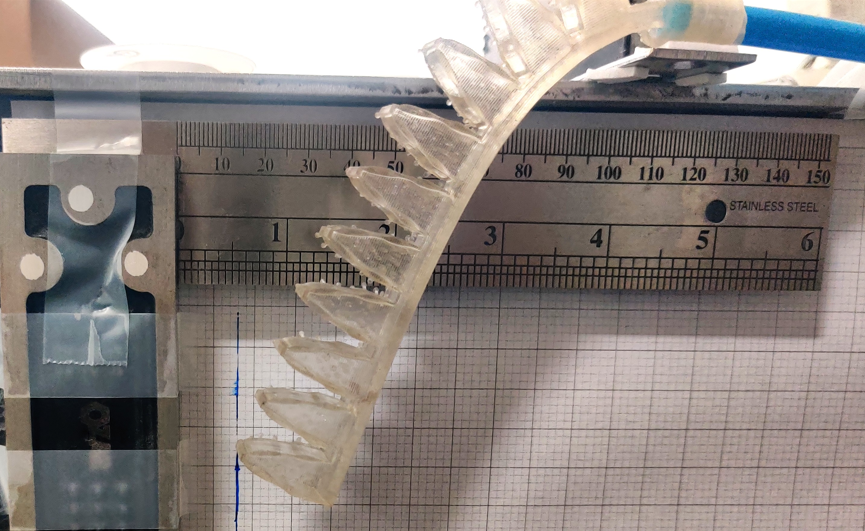}}           & \multicolumn{1}{c|}{19.11}  & \multicolumn{1}{c|}{\rule{0pt}{2.5cm}\includegraphics[scale=0.15]{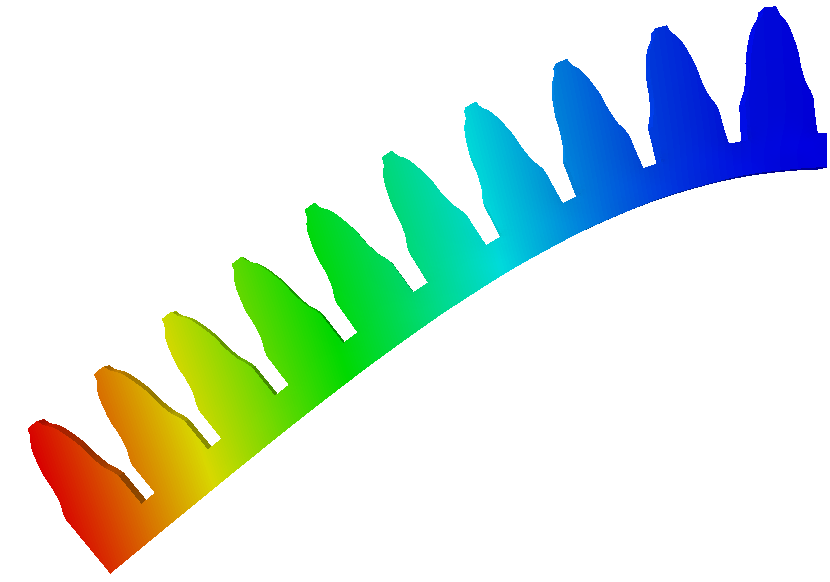}}    & 22.17 &13.80                             \\ \hline
		10                                                                                & \multicolumn{1}{c|}{\includegraphics[scale=0.15]{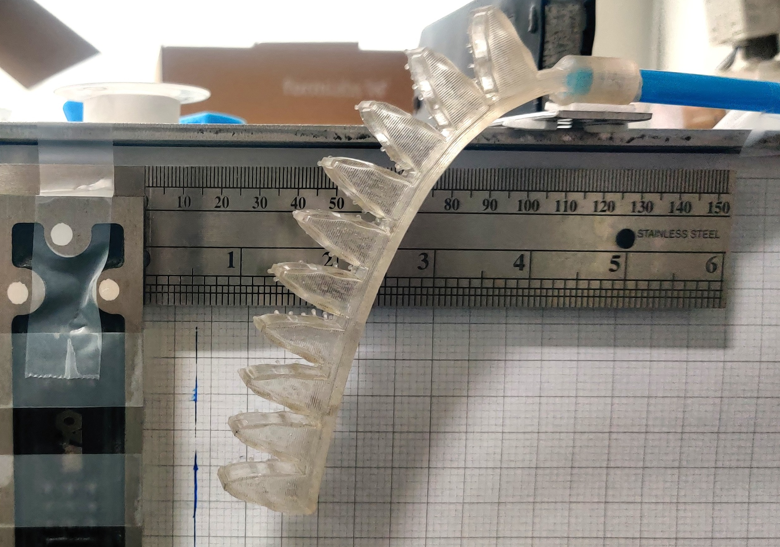}}           & \multicolumn{1}{c|}{26.70}  & \multicolumn{1}{c|}{\rule{0pt}{2.5cm}\includegraphics[scale=0.15]{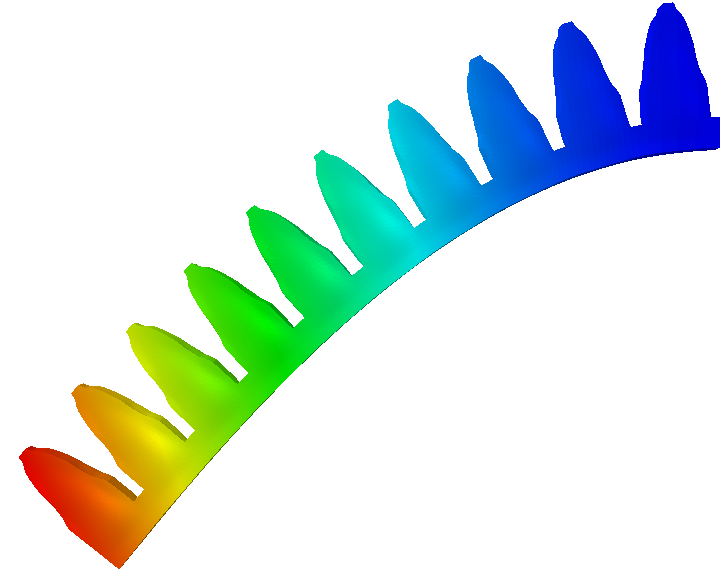}}    & 31.80  &    15.91                         \\ \hline
		20                                                                                & \multicolumn{1}{c|}{\includegraphics[scale=0.15]{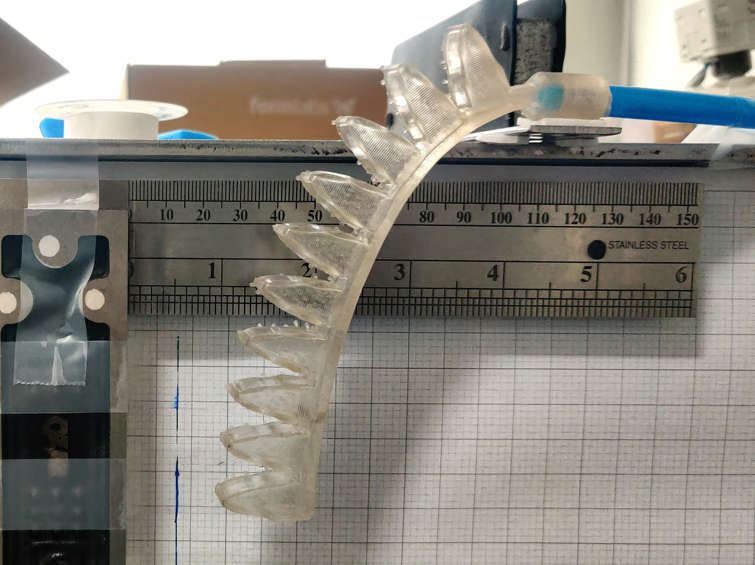}}           & \multicolumn{1}{c|}{35.43}  & \multicolumn{1}{c|}{\rule{0pt}{2.5cm}\includegraphics[scale=0.15]{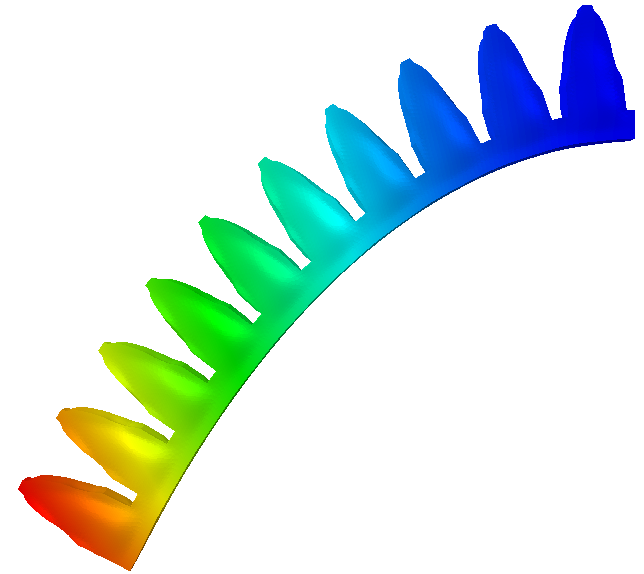}}    & 35.90  &    1.30                         \\ \hline
		32                                                                                & \multicolumn{1}{c|}{\includegraphics[scale=0.15]{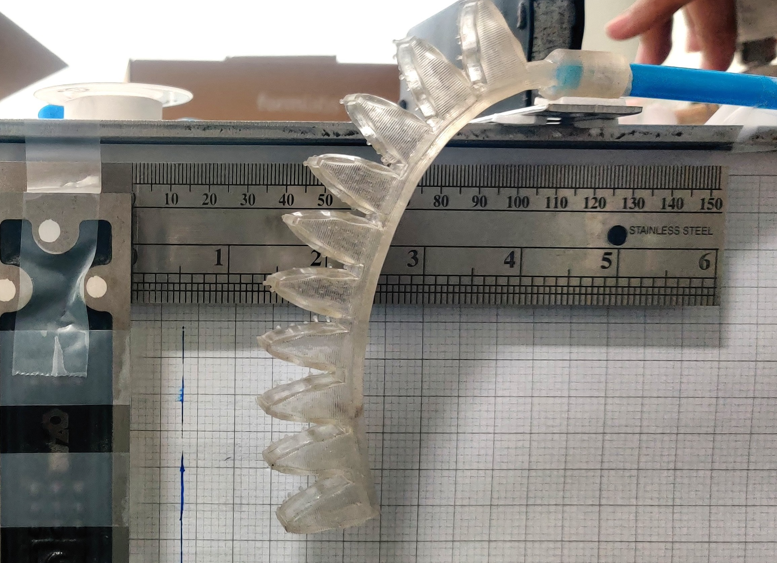}}           & \multicolumn{1}{c|}{46.71}  & \multicolumn{1}{c|}{\rule{0pt}{2.5cm}\includegraphics[scale=0.15]{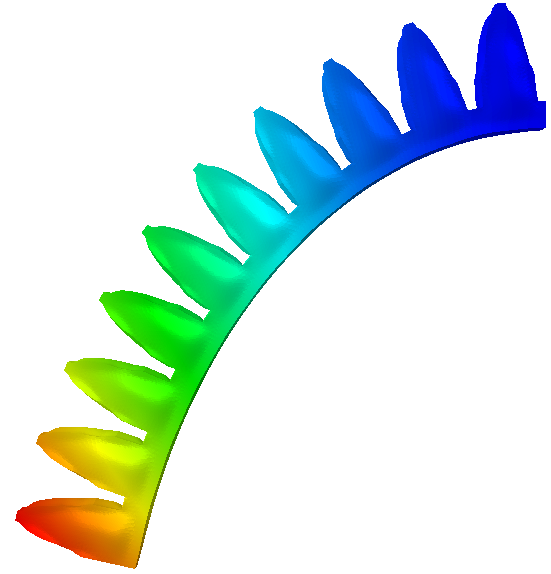}}    & 52.33 &    1.07                         \\ \hline
		40                                                                                & \multicolumn{1}{c|}{\includegraphics[scale=0.15]{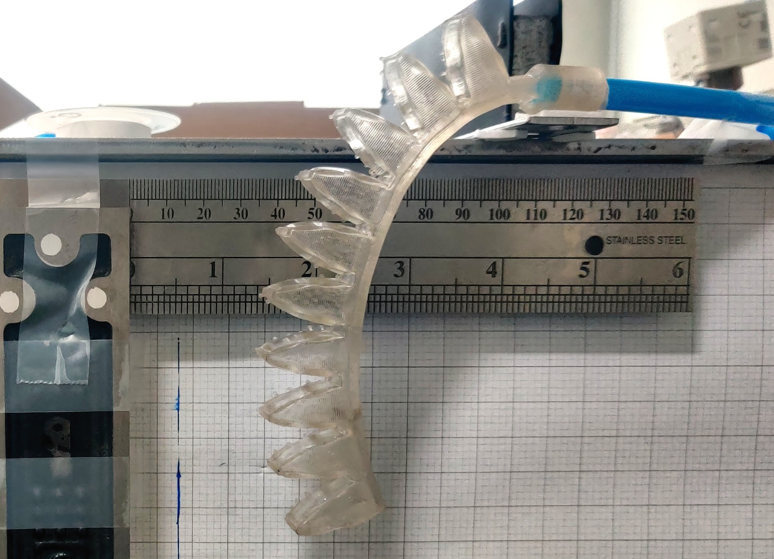}}           & \multicolumn{1}{c|}{54.73}  & \multicolumn{1}{c|}{\rule{0pt}{2.5cm}\includegraphics[scale=0.15]{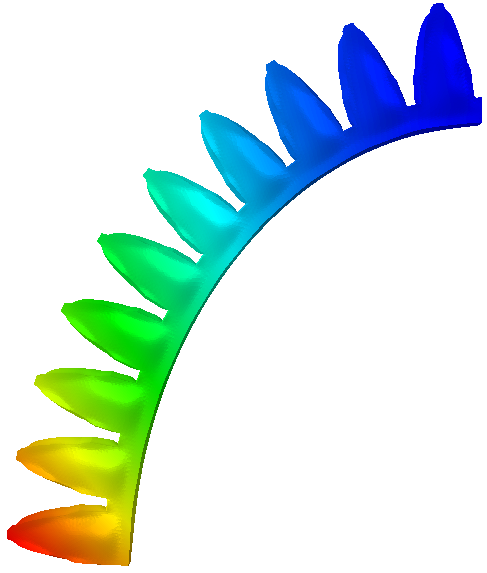}}    & 55.33 &      1.08                       \\ \hline
		52                                                                                & \multicolumn{1}{c|}{\includegraphics[scale=0.15]{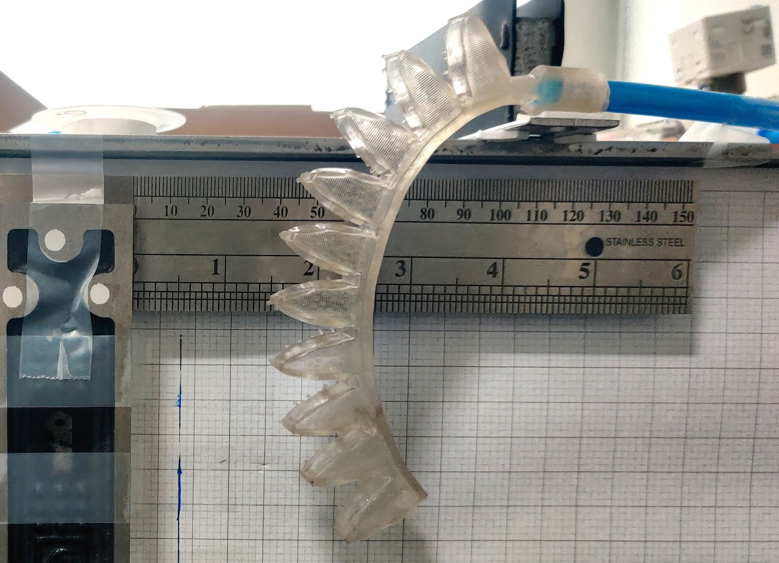}}           & \multicolumn{1}{c|}{67.42}  & \multicolumn{1}{c|}{\rule{0pt}{2.5cm}\includegraphics[scale=0.15]{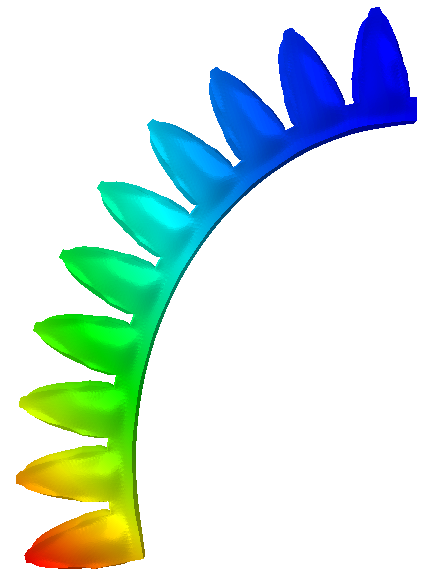}}    & 66.14  &     -1.94                        \\ \hline
		60                                                                                & \multicolumn{1}{c|}{\includegraphics[scale=0.15]{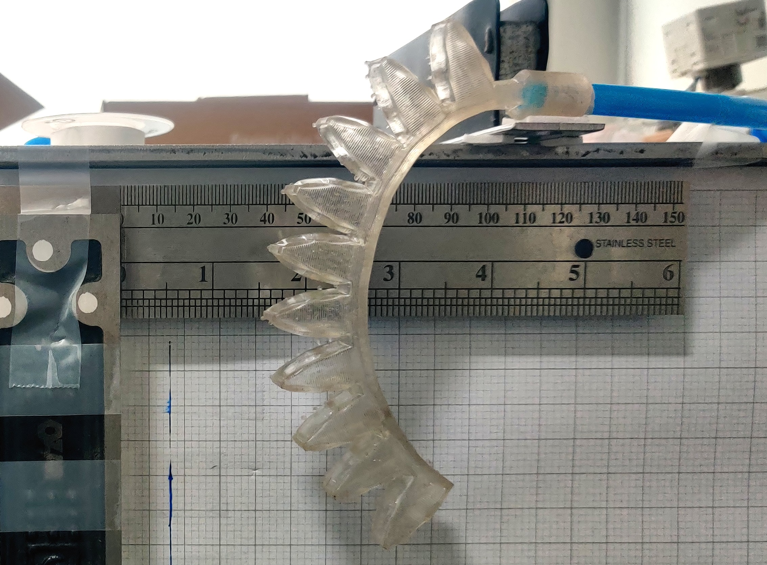}}           & \multicolumn{1}{c|}{68.18}  & \multicolumn{1}{c|}{\rule{0pt}{2.5cm}\includegraphics[scale=0.15]{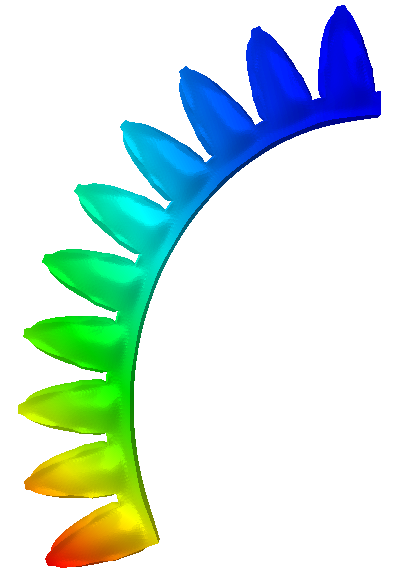}}    & 75.92  &     -0.34                        \\ \hline
	\end{tabular}
}
\end{table}

\subsection{3D printing and experimental demonstration}
This section provides 3D-printing process and experimental grasping action of the designed SPG.

\subsubsection{3D-printing process}
The optimized soft pneumatic gripper and its support structure are fabricated using the Formlabs Form 4 SLA 3D printer (Fig.~\ref{fig:3D-printer}). The stepwise procedure is described below:

\textbf{Design preparation:} The 3D model of the optimized soft pneumatic gripper is created using CAD software~(Fig.~\ref{fig:3D_CAD_opt}) and exported in STL format. 
\textbf{Printer and resin Selection:}
The components are printed using the Form 4 SLA printer (Fig.~\ref{fig:3D-printer}). Elastic 50A V2 Resin is used for the gripper to ensure flexibility and resilience, while  Grey Resin is used for the support structure due to its rigidity and dimensional stability.

\textbf{Slicing and setup:} The STL files are imported into \textit{PreForm} software. Support structures are auto-generated and manually adjusted for optimal print orientation and stability. The printer is loaded with the appropriate resin for each part.

\textbf{Printing process:} Printing is performed using SLA technology at a layer height of 100 microns, providing a balance between accuracy and build time. A precision laser selectively cures each layer of resin to form the parts.

\textbf{Post-processing:} After printing, the parts are rinsed in Form Wash (Fig.~\ref{fig:3D-printer}) using isopropyl alcohol (IPA) to remove excess uncured resin. They are then post-cured in Form Cure (Fig.~\ref{fig:3D-printer}) under UV light to finalize the mechanical properties and ensure structural integrity.

\textbf{Support removal and finishing:} Supports are carefully removed from the gripper and support structure to prevent damage. Minor finishing is done manually to ensure smooth surfaces and functional clarity.

\subsubsection{Comparison between FEA and experimental deformation}
A comparison between numerical and experimental deformations is shown in Table~\ref{Tab:3DdeformationNumExp}. We use air compressor with a pressure regulator to apply the pressure load. One notes that the discrepancies between the two sets of results are close to 1\%, except for outliers at $\SI{0}{\kilo\pascal}$ (only gravity load) and $\SI{10}{\kilo\pascal}$, which may have originated due to pre-stretched or residual stresses. Close agreement for the remaining loading conditions indicates that the selected material parameters for the Ogden material model are, for the most part, suitable for adequately capturing the nonlinear behavior of the arm with the applied pneumatic actuation.
\begin{figure}[H]
	\centering
	\begin{subfigure}[]{0.3\textwidth}
		\centering
		\includegraphics[scale=0.64]{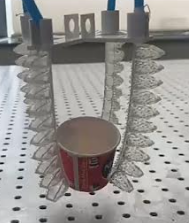}
		\caption{}
		\label{fig:ball_gripping}
	\end{subfigure}
	\quad
	\begin{subfigure}[]{0.3\textwidth}
		\centering
		\includegraphics[scale=0.75]{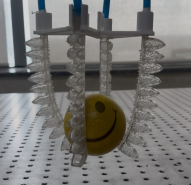}
		\caption{}
		\label{fig:cup_gripping}
	\end{subfigure}
	\quad
	\begin{subfigure}[]{0.3\textwidth}
		\centering
		\includegraphics[scale=0.655]{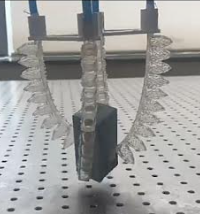}
		\caption{}
		\label{fig:chipbox_gripping}
	\end{subfigure}
	\begin{subfigure}[]{0.3\textwidth}
		\centering
		\includegraphics[scale=0.53]{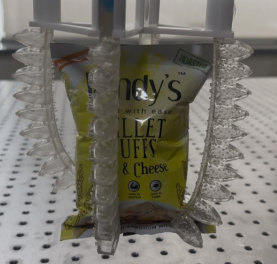}
		\caption{}
		\label{fig:rect_gripping}
	\end{subfigure}
	\quad
	\begin{subfigure}[]{0.3\textwidth}
		\centering
		\includegraphics[scale=0.45]{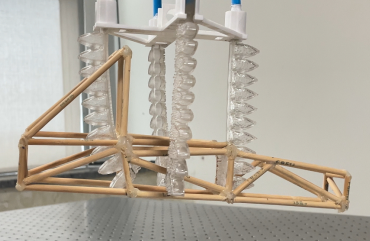}
		\caption{}
		\label{fig:cylinder_gripping}
	\end{subfigure}
	\quad
	\begin{subfigure}[]{0.2\textwidth}
		\centering
		\includegraphics[scale=0.29]{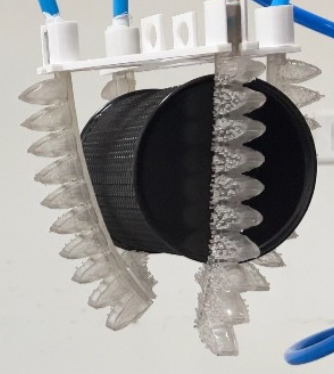}
		\caption{}
		\label{fig:chassis_gripping}
	\end{subfigure}
	\caption{Demonstration of the grasping capability of the proposed soft pneumatic gripper across a range of objects.} \label{fig:Expgripping}
\end{figure}

\begin{table}[H]
	\centering
	\caption{Required pneumatic pressure for gripping various objects}\label{Tab:GrippingP}
	\begin{tabular}{c|c|c ll}
		\cline{1-3}
		\textbf{Object}    & \textbf{Type} & \textbf{Gripping pressure}  (kPa)&  &  \\ \cline{1-3}
		Paper cup          & Fragile       & 13                                                                          &  &  \\ [0.5em] \cline{1-3}
		Soft toy ball      & Compliant     & 16                                                                          &  &  \\  [0.5em]\cline{1-3}
		Rectangular box    & Rigid         & 20                                                                          &  &  \\ [0.5em]\cline{1-3}
		Snack pouch        & Deformable    & 22                                                                          &  &  \\  [0.5em]\cline{1-3}
		Wooden Chassis       & Structural    & 28                                                                          &  &  \\  [0.5em]\cline{1-3}
		Cylindrical object & Metal         & 35                                                                          &  &  \\ \cline{1-3}
	\end{tabular}
\end{table}
\subsubsection{Gripping action demonstration}
The experimental setup is developed to demonstrate the gripping capability of the designed SPG. As shown in Fig.~\ref{fig:Expgripping}, the SPG comprises four arms mounted on a support structure that also contain the tube assembly for pneumatic load which is applied using an air compressor with a digital pressure regulator. To evaluate the real-world performance of the optimized and 3D-printed SPG, a series of functional tests are conducted using objects of varying shapes, sizes, weights, and stiffness. The test objects included a paper cup, a rectangular box, a soft ball, a snack packet, a wooden chassis, and a stationary holder. The gripping performance with these objects is illustrated in Fig.~\ref{fig:Expgripping} and required pressure load to grip the objects is depicted in Table~\ref{Tab:GrippingP}. These results demonstrate that the proposed SPG is capable of successfully handling objects with diverse geometries and material stiffness.

\section{Closing remarks}\label{Sec:Con}
A systematic topology optimization framework is proposed to design a soft pneumatic gripper while considering the design-dependent characteristics of the actuating loads. The approach automatically generates the optimized topology without requiring prior assumptions about its geometry. Since the task-performing behavior of the soft pneumatic gripper resembles that of a pressure-actuated compliant mechanism, the bending action of the arms is formulated as a min-max optimization problem with the output deformation of a 2D soft pneumatic unit. \PKadd{This formulation is developed} within a robust optimization framework \PKadd{in conjunction with linear deformation assumptions}. The blueprint and eroded designs are considered within the formulation. The volume constraint is imposed on the former, whereas a strain-energy constraint is applied to the latter design. The latter constraint ensures the optimized structure can withstand the applied pressure load. The 2D optimized soft pneumatic unit is then converted into a CAD model. The finite element results show that the optimized unit outperforms the conventional rectangular design at an applied pressure. The Ogden material model is used to capture material nonlinearity. 

The design-dependent nature of the actuating load is modeled using Darcy's law with an additional drainage term. The obtained pressure field is subsequently converted to nodal forces. Sensitivities of the objective and constraints are computed using the adjoint variable method. These quantities--objective, volume, and strain-energy constraints, along with their sensitivities--are then supplied to the MMA optimizer in its default min-max setting, ultimately yielding the optimized pressure chamber unit.

The CAD model of the optimized 2D Pneunet is extruded to generate the corresponding 3D unit, and ten such units are assembled to form a single arm of the presented SPG. Finite element analyses are conducted to demonstrate the bending behavior under varying pressure loads, both with and without the influence of self-weight. Results indicate that when gravity is considered, higher pressure is required to achieve bending motion, as expected, since gravitational forces oppose upward deformation. \PKadd{A comparative study shows that the arm with optimized chambers offers superior bending performance compared to that with conventional chambers, highlighting the effectiveness of the proposed approach.} Additionally, a comparison between numerical and experimental deformation profiles of the arm with optimized chambers, by and large, demonstrates good overall agreement, suggesting that the chosen Ogden material model parameters for nonlinear finite element analyses are suitable. The four arms are 3D-printed using stereolithography with Elastic 50A V2 resin and subsequently assembled to construct the SPG. An experimental setup is developed using the four arms and a supporting structure. The fabricated SPG demonstrates its grasping action by gripping and picking objects of varying shapes, sizes, stiffness, structures, and weights at different pneumatic loads.

  \section*{Declaration of Competing Interest}
  The authors declare that they have no known competing financial interests or personal relationships that could have appeared to influence the work reported in this paper.
  \section*{Acknowledgment}
  PK thanks  Anusandhan National Research Foundation, India for the support under MATRICS project file number MTR/2023/000524. The authors thank Amal Shaji and Nikhil Chavan for their assistance with the experimental setup and are grateful to Krister Svanberg for providing the MATLAB implementation of the Method of Moving Asymptotes used in this work.

\end{document}